\documentclass[a4paper, fleqn]{cas-sc}

\usepackage[numbers]{natbib}

\def\tsc#1{\csdef{#1}{\textsc{\lowercase{#1}}\xspace}}
\tsc{WGM}
\tsc{QE}

\usepackage{amsmath,amssymb,bm}
\usepackage{mathtools}
\usepackage{booktabs}
\usepackage{xspace}

\usepackage{enumitem}
\usepackage{algorithm}
\usepackage{algpseudocode}
\usepackage[table]{xcolor}

\newcommand{\R}{\mathbb{R}}

\begin{document}
  \let\WriteBookmarks\relax \def\floatpagepagefraction{1} \def\textpagefraction{.001}

  \shorttitle{Distributed Predictive Flocking for a Swarm of Miniature Vibration-Driven Robots in a Confined Arena}

  \shortauthors{Shrestha \textit{et al.}}

  \title[mode = title]{Distributed Predictive Flocking for a Swarm of Miniature Vibration-Driven Robots in a Confined Arena}

  \tnotemark[1]

  \tnotetext[1]{This work was supported by the Pattern Recognition Lab, Friedrich-Alexander-Universit\"{a}t Erlangen-N\"{u}rnberg.}

  \author[1]{Prajol Shrestha}[orcid=0009-0005-9585-3800]
  \cormark[1]
  \ead{prajol.shrestha@fau.de}
  \credit{Conceptualization, Methodology, Software, Formal Analysis, Writing -- Original Draft, Writing -- Review \& Editing, Visualization, Validation}

  \author[2]{Siddhant Mohapatra}
    \credit{Conceptualization, Formal Analysis, Visualization, Writing -- Review \& Editing}
  \author[2]{Filip Novkoski}
      \credit{Formal Analysis, Visualization, Writing -- Review \& Editing}
  \author[1]{Julian Oelhaf}
    \credit{Writing -- Review \& Editing}
  \author[3]{Nicolas Vandewalle}
      \credit{Writing -- Review \& Editing}
  \author[2,4]{Ana-Sun\v{c}ana Smith}
    \credit{Writing -- Review \& Editing}
  \author[1]{Andreas Maier}
  \ead{andreas.maier@fau.de}
  \credit{Supervision, Resources, Project administration, Funding acquisition, Conceptualization, Writing -- Review \& Editing}
  
  \affiliation[1]{organization={Pattern Recognition Lab, Department of Computer Science, Friedrich-Alexander-Universit\"{a}t Erlangen-N\"{u}rnberg}, postcode={91058}, city={Erlangen}, country={Germany}}
  \affiliation[2]{organization={PULS, Institute for Theoretical Physics, Friedrich-Alexander-Universit\"{a}t Erlangen-N\"{u}rnberg}, postcode={91058}, city={Erlangen}, country={Germany}}
  \affiliation[3]{organization={GRASP, Institute of Physics B5a, University of Li\`{e}ge}, postcode={B4000}, city={Li\`{e}ge}, country={Belgium}}
  \affiliation[4]{organization={Group for Computational Life Sciences, Division of Physical Chemistry, Ru\dj{}er Bo\v{s}kovi\'{c} Institute}, postcode={10000}, city={Zagreb}, country={Croatia}}

  \cortext[1]{Corresponding author}



  \begin{abstract}
    Coordinating many miniature robots in a confined arena, with no leader, central planner, or global map, is a core challenge for swarm robotics. As the swarm grows denser, collision avoidance becomes the limiting factor, and the number of safely coexisting robots depends on the formulation of the inter-robot safety constraint. We present a distributed predictive flocking controller in which each robot solves its own nonlinear model predictive control (NMPC) problem, optimizing swarm cohesion and alignment costs, inter-robot and obstacle safety, and a task objective, while exchanging only predicted trajectories with local neighbors. Within this generalized controller architecture, solved in real time, we compare six barrier and control-barrier-function (CBF) safety formulations on a digital twin of an open-source vibration-driven bristlebot in a $0.64\,\mathrm{m^2}$ arena. A look-ahead discrete-time CBF performs best overall, with the fewest collisions and the widest inter-robot margins across nearly every swarm size tested. We deploy it on phototaxis, phototactic orbiting, and cohesion without a stimulus tasks, using swarm sizes starting from ten robots and up to the geometric packing limit. The swarm holds well: collisions stay low in small swarms, rising steeply as the packing limit approaches, while coordination and cohesion degrade gently rather than collapsing. The swarm is found not to trade safety for order, keeping both across the confined arena, with local alignment high at every swarm size.
  \end{abstract}


  \begin{highlights}
    \item Distributed leader-free predictive flocking for a dense swarm of miniature robots.
    \item Six safety formulations compared on a single common controller footing.
    \item Look-ahead safety constraint gives widest margins and fewest collisions.
    \item Single generalized controller deployed to handle different tasks successfully.
    \item  Swarm stays coordinated up to the geometric packing limit in each task.
  \end{highlights}


  \maketitle

\section{Introduction}
\label{sec:introduction}

Autonomous multi-robot systems are a promising approach for tasks in cluttered,
inaccessible, or hazardous spaces, from confined-structure inspection to search in
collapsed buildings and medical micro-robotics~\citep{yang2018grand}. In these settings,
many miniature robots must coordinate in proximity within a tight, bounded arena. As
in natural swarms, such as flocking of starlings, schooling of fish, the crowded colonies
of ants and honey bees, and even the collective migration of cells, robust group behavior
emerges from purely local interactions among simple
agents~\citep{camazine2001self,vicsek1995novel,jadbabaie2003coordination}, with no central
control~\citep{yang2018grand}.
Miniaturization, however, tightly bounds onboard compute, memory, and
power~\citep{neuman2022tiny}, demanding control that is theoretically principled yet
computationally light. Moreover, when the arena is fixed, a growing number of robots
increases the density of the swarm and hinders smooth motion. When it is also cluttered
with obstacles, coordination is even harder. In this scenario, collision avoidance becomes
the central challenge.

The standard way to coordinate such a swarm from purely local interactions is Reynolds'
boid model~\citep{reynolds1987flocks,reynolds1999steering}, which reduces collective
motion to three local rules (cohesion, alignment, and separation) that produce coherent
group motion without global information~\citep{couzin2002collective}. Early control laws
formalized these as artificial potential fields over a communication
graph~\citep{olfati2006flocking,tanner2003stable}, and reactive
velocity-matching rules scale to tens of drones~\citep{vasarhelyi2018optimized}. Such
reactive methods, however, offer no predictive horizon, no formal safety guarantee under
nonlinear dynamics, and no systematic handling of kinodynamic limits. These shortcomings
are especially severe in confined, high-density settings, where many underactuated robots
have little room to recover from unsafe motions.

Model predictive control (MPC)~\citep{rawlings2017model} addresses these limitations. For
the nonlinear dynamics of mobile robots, this takes the form of nonlinear MPC (NMPC). It
solves a
finite-horizon optimal control problem (OCP) online, which provides closed-loop feedback,
constraint satisfaction, and explicit multi-objective trade-offs. Distributed MPC (DMPC)
extends this to a swarm by splitting the
global problem into per-agent OCPs coupled through shared
constraints~\citep{van2017distributed,luis2020online,testa2021choirbot,pichierri2023crazychoir}.
Solvers such as the Sequential Quadratic Programming real-time iteration
(SQP-RTI)~\citep{verschueren2022acados} solve each per-agent OCP in real time. Similarly,
optimization-based
swarm planners achieve strong real-world navigation~\citep{zhou2022swarm,
10599811, 11128011, csenbacslar2024dream}. These methods are mainly developed for
point-to-point navigation~\citep{luis2020online,9851518}, with collision avoidance
typically encoded as geometric constraints such as separating hyperplanes or safe
corridors.

The closest works embed Reynolds behaviors in NMPC. They have been demonstrated on up to 16
quadrotors~\citep{soria2021predictive,soria2021distributed}, leader-follower
teams~\citep{nag2022flock,Nag18052024,10.3389/frobt.2025.1540808}, and fixed-wing unmanned
aerial vehicles (UAVs)~\citep{10240689}. Most share one regime. They assume sparse flight in large 3D
airspace, with the swarm migrating toward a shared goal. Even when several obstacles are
present, they remain sparse relative to the available space, keeping the environment
effectively low-density. Few address the opposite regime that we target, where many
underactuated robots coordinate at high density in a fixed, confined, and cluttered
arena under a task objective such as following a light gradient. Here, the
inter-robot safety formulation, not free-space planning, governs how many robots can
coexist. Each of these works, however, fixes a single safety formulation in advance,
usually a distance constraint or a penalty. It remains open which formulation fits the
most robots into a fixed arena with minimal collisions.

Safety formulations differ in how collision avoidance is encoded as a constraint in the
OCP. Such a constraint is classically formulated using a barrier function $h(x)$ whose safe set $\mathcal{S} = \{x \mid h(x)\ge 0\}$ is kept forward invariant to certify safety~\citep{ames2016control}, typically imposed as a
hard~\citep{salimi2021nmpc} or soft~\citep{lindqvist2020nmpc} distance constraint. A Control
Barrier Function (CBF)~\citep{ames2016control} relaxes this strict boundary condition by allowing $h(x)$ to decrease at a rate bounded by a class-$\mathcal{K}$ function, and its discrete-time form
(Dt-CBF)~\citep{agrawal2017discrete} imposes this at each MPC stage.
\citet{zeng2021mpccbf} enforces it as a hard constraint, which is often infeasible.
Feasibility is restored by softening the decay with slack
variables~\citep{zeng2021feasibility,zeng2021optimaldecay,nguyen2024relaxedcbf}, adding
terminal barrier terms~\citep{wabersich2023predcbf,katriniok2023dtcbf}, or anchoring a
soft CBF with one hard step~\citep{lu2024safeplanning}. Beyond feasibility, when the constraint
has relative degree greater than one, as for a distance constraint under unicycle
dynamics, a higher-order CBF (HOCBF)~\citep{xiao2019hocbf,nguyen2016exponential} chains
the barrier conditions. \citet{lu2024safeplanning} compare a distance-constraint MPC, a
hard discrete-time CBF, a slacked CBF, and their combination on a common footing, but for
a \emph{single} robot. In the multi-robot setting, safety is enforced by reactive
schemes~\citep{vandenberg2011reciprocal,zhou2017buffered}, by post-hoc CBF-QP (quadratic
program) filters~\citep{wang2017barriercerts}, or by a CBF inside a decentralized MPC with
neighbors as predicted
trajectories~\citep{mali2021predictioncbf,wang2025dsmpc,chandra2025deadlock,goarin2025decentralized}.
A formulation's collision performance, however, likely changes with the swarm density. A
comparison across densities is required to find the best one.

In this work, we investigate the maximum number of miniature robots a distributed
predictive controller can flock in a fixed arena, and design a controller to push that
limit. Each robot solves its own NMPC combining Reynolds cohesion, alignment, and
separation (gated by neighbor presence) with inter-robot/obstacle safety and the task
objective. It selects the $K$ nearest neighbors within the communication radius and
exchanges predicted trajectories synchronously as OCP parameters, giving a fully
distributed, leader-free architecture with no central planner or global map. We evaluate
it on a physics-informed digital twin~\citep{rd1t-h2v6} of the
GRASPion~\citep{10.1063/5.0303152}, an open-source vibration-driven bristlebot studied as
an active-matter model system. We characterize how densely the swarm packs a
$0.64\,\mathrm{m^2}$ arena while sustaining two behaviors. These are \emph{phototaxis} (ascending toward a
moving Gaussian light gradient) and a \emph{phototactic orbital} mode (orbiting the source
along the intensity level-set). We compare six inter-robot safety formulations on a common
footing (Section~\ref{sec:comparison}). A slacked barrier paired with a discrete-time CBF
at a forward handle~\citep{huang2023unicyclecbf} sustains the highest density with the
fewest collisions.
The main contributions of this paper are as follows:
\begin{enumerate}
  \item A distributed predictive control system for dense swarms of miniature vibration-driven robots. Each robot solves its own NMPC fusing flocking, safety, and the task objective, coupled to neighbors only through exchanged predicted trajectories.

  \item Demonstration that a single controller generalizes across qualitatively different collective behaviors. The same architecture produces gradient-ascending phototaxis, stable phototactic orbital formation, and generic flocking with no modification to the safety or coordination layers, showing that task objectives and coordination structure are separable and that new tasks can be added without re-deriving safety guarantees.

  \item A careful comparison of six inter-robot safety formulations within one identical controller, isolating the safety constraint as the only varying factor. The Look-ahead Slack Dt-CBF gives the best overall trade-off for the phototaxis deployment target.

  \item Characterization of the operating swarm density that the controller sustains in a fixed arena. The swarm stays cohesive with few collisions in small swarms; collisions rise steeply towards the geometric packing limit, as saturation sets in, while coordination and feasibility degrade gently rather than collapse.
\end{enumerate}

The remainder of this paper is structured as follows. Section~\ref{sec:NMPC} presents the robot model, light field, control architecture overview, behavior generation, flocking rules, and the DMPC OCP. Section~\ref{sec:simulation} reports the simulation results, including the safety-formulation study (Section~\ref{sec:comparison}), the analysis of collective behavior (Section~\ref{sec:collective_behavior}), and the quantification of collision avoidance under different tasks (Section~\ref{sec:density_comparison}). Section~\ref{sec:conclusion} summarizes the work and provides an outlook for future studies.

\section{Preliminaries and Problem Statement}
\label{sec:NMPC}

In this work, we consider a swarm of $N_b$ robots operating in a bounded arena
$\mathcal{A} = [-A_h,\,A_h]^2 \subset \mathbb{R}^2$, with each robot labeled by
$i \in \{1,\ldots,N_b\}$. The state of robot $i$ is
$\bm{x}_i = [\bm{p}_i,\,\theta_i]^\top \in \mathbb{R}^3$, where
$\bm{p}_i = [p_{x,i},\,p_{y,i}]^\top$ is the position and $\theta_i$ the heading.
Inter-robot communication is modeled as a directed sensing graph
$\mathcal{G} = (\mathcal{V},\,\mathcal{E})$, where
$\mathcal{V} = \{1,\ldots,N_b\}$ denotes the set of robots and
$\mathcal{E} \subseteq \mathcal{V} \times \mathcal{V}$ is the communication set of ordered pairs $(i,j)$ for which robot $i$ can observe robot $j$. The neighbor set of robot $i$ is $\mathcal{N}_i = \{j \in \mathcal{V} \mid (i,j) \in \mathcal{E}\}$. We adopt a topological neighborhood with fixed cardinality
$|\mathcal{N}_i| \le K$, selecting the $K$ nearest robots within communication radius $R_{\rm comm}$ of robot $i$. This choice keeps the neighbor count constant and avoids the cost discontinuities that arise under Voronoi or attraction-based topologies~\citep{ballerini2008interaction, strandburg2010visual, chen2013multi}. Each robot additionally accounts for up to $M$ nearby static circular obstacles within a sensing range $d_{\rm obs}$, indexed by $m \in \{1,\ldots, M\}$, with obstacle $m$ at center $\bm{p}_m \in \mathbb{R}^2$ and radius $r_m$. The environment also includes a single light source whose intensity field serves as the swarm stimulus. 

The OCP is compiled once with a fixed parameter dimension. Neighbor and obstacle
information therefore enters two fixed-size blocks of $K$ and $M$ slots, the neighbor block and the obstacle block. At each control step, robot $i$ receives a predicted trajectory $\bm{\tau}_j = \bigl\{[\hat{\bm{p}}_j^{(k)},\,\hat{\bm{v}}_j^{(k)}]^\top\bigr\}_{k=0}^{N}$
from each neighbor $j \in \mathcal{N}_i$, where $\hat{\bm{p}}_j^{(k)} \in \mathbb{R}^2$ and $\hat{\bm{v}}_j^{(k)} \in \mathbb{R}^2$ are the predicted position and velocity of neighbor $j$ at horizon stage $k$. Throughout, a hat denotes a quantity the per-robot OCP treats as fixed data (the ego robot's current measured state and the neighbors' received predicted trajectories). The ego robot's own predicted states are decision variables. These trajectories fill the neighbor block, and the number of occupied slots changes from step to step. Each neighbor slot therefore carries a binary activation mask $a_j^{(k)} \in \{0,1\}$, where $a_j^{(k)} = 1$ if slot $j$ is occupied at stage $k$ and $a_j^{(k)} = 0$ otherwise. The obstacle block is filled the same way. Robot $i$ selects the $M$ nearest obstacles within obstacle sensing range $d_{\rm obs}$ and writes each center $\bm{p}_m$ into a slot together with its own activation flag.

\subsection{Robot Model}
\label{sec:robot_model}
\begin{figure}[pos=htbp]
  \centering
  \begin{minipage}{0.48\textwidth}
    \centering
    \includegraphics[width=\linewidth]{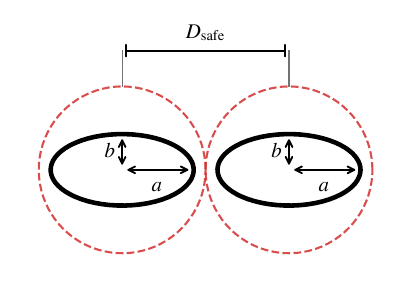}
  \end{minipage}\hfill
  \begin{minipage}{0.48\textwidth}
    \centering
    \includegraphics[width=\linewidth]{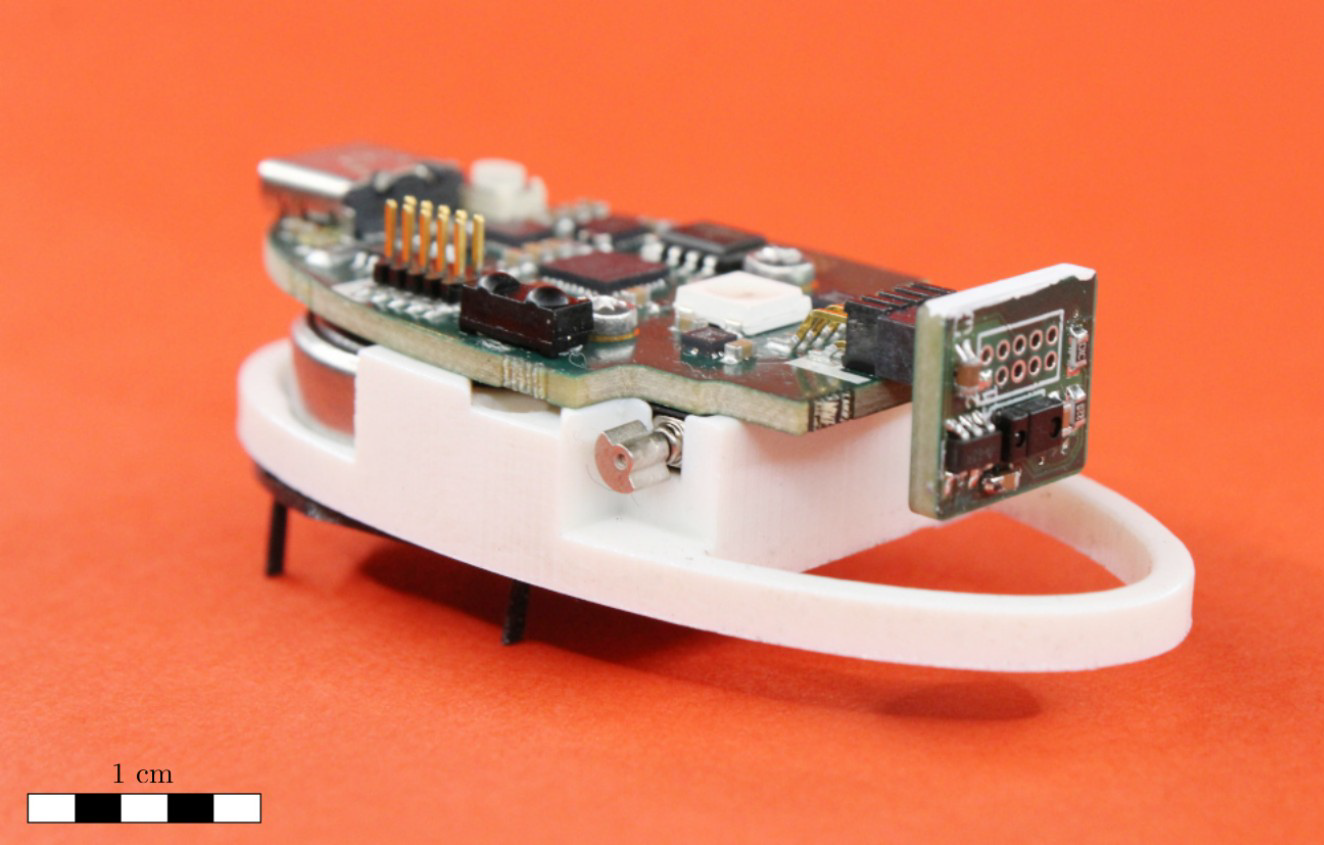}
  \end{minipage}
  \caption{(Left) Two elliptical robots with minimum safe distance $D_{\rm safe}$ and semi-axes $a$ ($3\,\mathrm{cm}$) and $b$ ($1.5\,\mathrm{cm}$). (Right) The physical GRASPion open-source bristlebot hardware.}
  \label{fig:robot_model}
\end{figure}

Each robot is modeled as a rigid ellipse with semi-major axis $a$ and semi-minor axis $b$,
matching the physical footprint of the GRASPion open-source bristlebot~\citep{10.1063/5.0303152}, as illustrated in Fig.~\ref{fig:robot_model}. The
simulated elliptical agent is modeled after this hardware platform, following the
physics-informed digital twin of~\citep{rd1t-h2v6}, so the simulation study is grounded in
a physical robot footprint and can serve as a precursor to future physical experiments.

The dynamics of each robot are described by a three-state kinematic unicycle model with state $\bm{x} \in \mathbb{R}^3$, input $\bm{u} \in \mathbb{R}^2$, and continuous-time dynamics $f(\bm{x}, \bm{u})$ given by
\begin{align}
  \bm{x} &= [\,p_x,\, p_y,\, \theta\,]^\top, \label{eq:state}\\
  \bm{u} &= [\,v,\, \omega\,]^\top, \label{eq:input}\\
  \dot{\bm{x}} &= f(\bm{x},\bm{u}) = [\,v_x,\, v_y,\, \omega\,]^\top, \label{eq:unicycle}
\end{align}

where $(p_x, p_y)$ is the position, $\theta$ the heading angle, $v$ the linear velocity, $v_x = v\cos\theta$ and $v_y = v\sin\theta$ are the velocity components, and $\omega$ the angular velocity. The continuous-time dynamics~\eqref{eq:unicycle} are integrated using an explicit Runge-Kutta 4 (ERK-4) integrator with three steps per control interval, yielding the discrete-time map $\bm{x}^{(k+1)} = F(\bm{x}^{(k)}, \bm{u}^{(k)})$.

\subsection{Light Field Model}
\label{sec:light_field}
The arena contains a time-varying light field that each robot senses locally to compute its navigation reference. The field is modeled as a Gaussian intensity profile whose source either orbits the arena center or remains stationary.

\paragraph{Light Source}
The light source produces a Gaussian intensity field
\begin{equation}
  I(\bm{p},t) = I_0\,\exp\!\left(-\frac{\|\bm{p} - \bm{s}(t)\|^2}{2\sigma_{\rm L}^2}\right),
  \label{eq:gaussian}
\end{equation}
where $\bm{p} \in \R^2$ is an arbitrary position, $I_0 > 0$ is the peak intensity, $\bm{s}(t) \in \R^2$ is the position of the source at time $t$, and $\sigma_{\rm L}$ represents the spatial spread of intensity.

\paragraph{Source Dynamics}
The light source supports two motion modes. In circular rotation mode, the source orbits the center of the arena at a fixed radius $r$ with angular speed $\omega_s = v_s /r$, with the position of the source given by
\begin{equation}
  \bm{s}(t) = r\begin{bmatrix}\cos(\omega_s t + \varphi_0)\\\sin(\omega_s t + \varphi_0)\end{bmatrix},
  \label{eq:rotation}
\end{equation}
where $\varphi_0$ is the initial angle of the source in the orbit and $v_s$ is the speed of the source along the orbit. In stationary mode, the source remains fixed at the arena center.

\paragraph{Intensity Gradient}
The intensity gradient with respect to the robot position is computed analytically as
\begin{equation}
  \nabla_{\bm{p}} I(\bm{p},t) =
  -\frac{(\bm{p} - \bm{s}(t))}{\sigma_{\rm L}^2}\,I(\bm{p},t).
  \label{eq:gradient}
\end{equation}
This analytic gradient is the local cue that the collective-behavior modes of
Section~\ref{sec:navigation} convert into each robot's look-ahead reference.

\subsection{Control Architecture Overview}
\label{sec:overview}

\begin{figure}[pos=htbp]
  \centering
  \includegraphics[width=\linewidth]{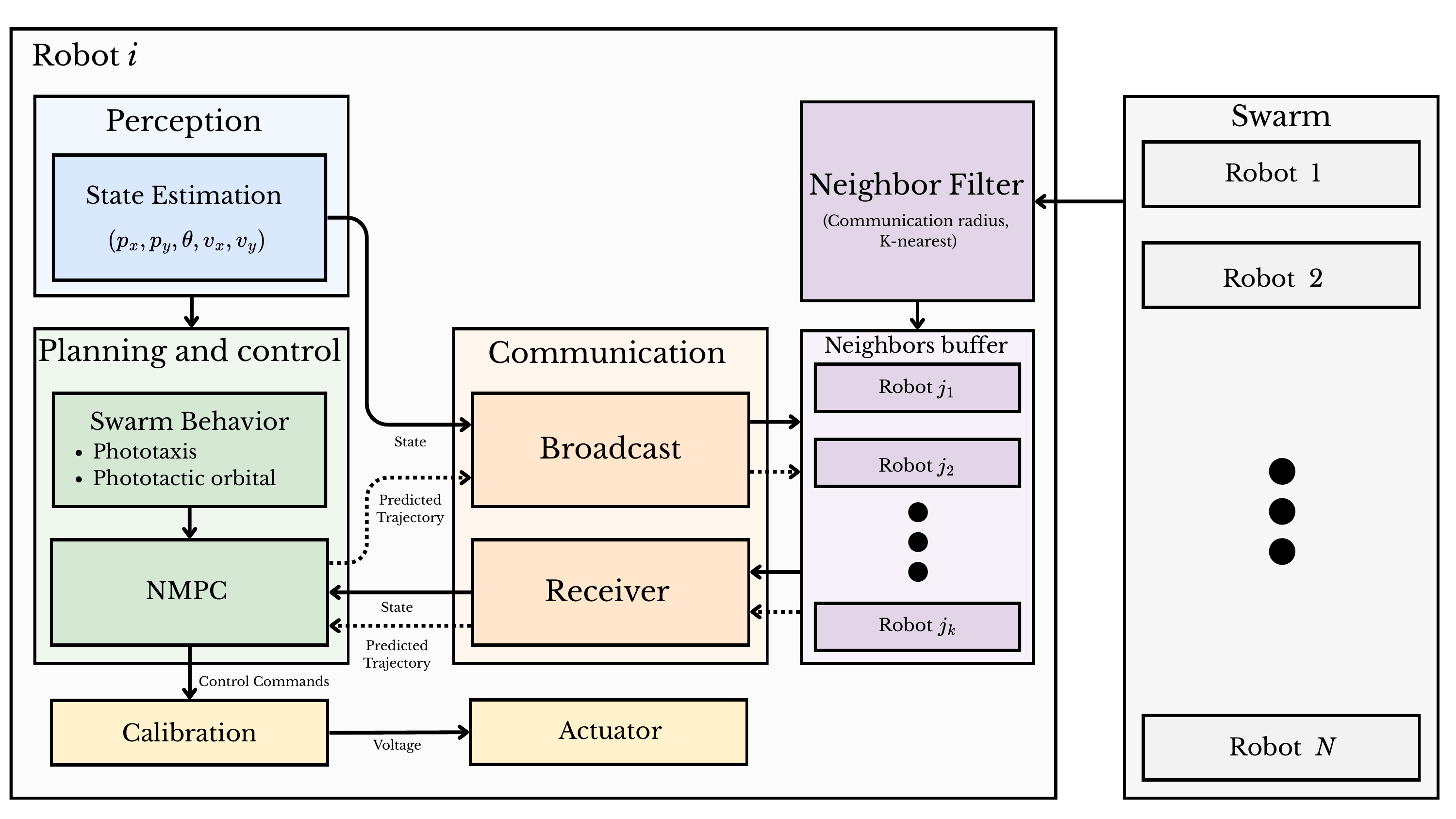}
  \caption{
  Per-robot control architecture for the DMPC swarm. Each robot runs three concurrent subsystems. The \emph{Perception} block estimates the robot's state comprising position $p_x$ and $p_y$, orientation $\theta$, and velocity $v_x$ and $v_y$. The \emph{Planning and Control} block uses this state to compute a
  look-ahead reference based on the active swarm behavior (phototaxis or
  phototactic orbital) and then solves the local NMPC problem to produce
  control commands, which are passed through a Calibration layer before
  reaching the Actuator. The \emph{Communication} block handles all inter-robot data exchange, where solid arrows denote state signals and dotted arrows denote
  predicted trajectory exchanges. Incoming broadcasts are filtered by a Neighbor Filter (within communication range $R_{\rm comm}$ and $K$-nearest selection) and stored in the robot's neighbors buffer before being used by the NMPC solver.}
  \label{fig:architecture}
\end{figure}

Every robot runs an identical decentralized controller, outlined in
Fig.~\ref{fig:architecture}. At each control step, the robot senses the local light field, and a behavior module maps this reading to a look-ahead reference that encodes where the robot should head next. A per-robot optimal control problem (OCP) tracks that reference. Flocking rules keep the group cohesive and heading-aligned, while barrier and control-barrier-function constraints enforce inter-robot and obstacle safety. Robots publish their predicted trajectories over the sensing graph and re-solve synchronously from their neighbors' previous plans, coordinating the swarm without a central node. The rest of this section develops the remaining components in turn. These are the behavior module
(Section~\ref{sec:navigation}), the flocking rules (Section~\ref{sec:reynolds}), the six safety formulations (Sections~\ref{sec:separation}--\ref{sec:obstacle_avoidance}), the reference-tracking and control-effort costs (Section~\ref{sec:la_cost}), the assembled OCP (Section~\ref{sec:ocp_full}), and the decentralized execution loop (Section~\ref{sec:sim_workflow}).

\subsection{Collective Behavior Modes}
\label{sec:navigation}

Collective motion emerges when individual agents produce coordinated group-level movement through purely local interactions in the absence of centralized
planning~\citep{reynolds1987flocks, vicsek2012collective}. In our controller, a behavior module translates local sensory observations into a look-ahead reference
$(\bm{p}^{\rm ref}, \theta^{\rm ref}, v^{\rm ref}, \omega^{\rm ref})$ at each control step, which the OCP then tracks while maintaining flocking coherence and safety. We describe two such modes, phototaxis and phototactic orbital behavior.

At each control step, the light-field gradient $\bm{g}=\nabla I(\hat{\bm{p}}_i)$ is evaluated analytically from~\eqref{eq:gaussian}--\eqref{eq:gradient}. The resulting reference heading depends on the selected collective behavior mode. Once the heading is obtained, the look-ahead reference position is computed as
\begin{equation}
  \bm{p}^{\rm ref}
  =
  \operatorname{clip}\!\left(
  \hat{\bm{p}}_i
  +
  d_{\rm la}\,[\cos\theta^{\rm ref},\,\sin\theta^{\rm ref}]^\top,
  [-A_h,A_h]^2
  \right)
  \label{eq:shared_ref}
\end{equation}
where $d_{\rm la}$ is the look-ahead distance, and the clipping function constrains the reference position to remain inside the arena. The reference velocity is set to a constant target value $v^{\rm ref}$, and the reference angular velocity $\omega^{\rm ref}$ is fixed at zero.

\subsubsection{Phototaxis Behavior Mode}
\label{sec:phototaxis}

Phototaxis drives each robot to ascend the intensity gradient of the light field without knowledge of the global source position. The reference heading is
aligned with the steepest ascent direction:
\begin{equation}
  \theta^{\rm ref} =
  \begin{cases}
    \operatorname{atan2}(g_y, g_x), & \|\bm{g}\|_2 \ge \epsilon_{\rm grad},\\
    \hat{\theta}_i, & \|\bm{g}\|_2 < \epsilon_{\rm grad},
  \end{cases}
  \label{eq:phototaxis_heading}
\end{equation}
where $\epsilon_{\rm grad}$ is a small gradient threshold. When the field is nearly flat ($\|\bm{g}\|_2 < \epsilon_{\rm grad}$), the robot holds its current heading $\hat{\theta}_i$ rather than turning randomly.

\subsubsection{Phototactic Orbital Behavior Mode}
\label{sec:orbit}

In the phototactic orbital mode, each robot follows the tangent of the light intensity level set, effectively orbiting the source without knowledge of its global position. The reference heading is obtained by rotating the gradient direction from~\eqref{eq:phototaxis_heading} by $\pi/2$, with the orbital direction governed by the circulation parameter $\kappa \in \{-1, +1\}$  (where $\kappa = +1$ generates counter-clockwise motion, and $\kappa = -1$ generates clockwise motion):
\begin{equation}
  \theta^{\rm ref} =
  \begin{cases}
    \operatorname{atan2}(g_y, g_x) - \kappa\frac{\pi}{2}, & \|\bm{g}\|_2 \ge \epsilon_{\rm grad},\\
    \hat{\theta}_i, & \|\bm{g}\|_2 < \epsilon_{\rm grad}.
  \end{cases}
  \label{eq:orbit_ref}
\end{equation}

\subsubsection{Extensibility to Arbitrary Behaviors}
\label{sec:extensibility}

Decoupling the reference generator from the OCP makes the architecture task-agnostic. Any behavior (directed migration, dispersal, transport, shape formation) can be integrated by mapping local observations to a look-ahead reference $(\bm{p}^{\rm ref}, \theta^{\rm ref}, v^{\rm ref}, \omega^{\rm ref})$, while the DMPC preserves safety and flocking coherence.

\subsection{Reynolds Flocking Rules}
\label{sec:reynolds}

The behavior modes set each robot's individual objective, while the collective motion arises from the local steering rules proposed by Reynolds~\citep{reynolds1987flocks, reynolds1999steering}. Cohesion binds robots into a group, alignment coordinates their heading, separation keeps them from colliding, and obstacle avoidance steers them clear of static hazards. The controller encodes cohesion and alignment as quadratic cost terms in the per-robot OCP, while separation and obstacle avoidance are enforced through barrier and discrete-time CBF constraints~\citep{ames2016control, agrawal2017discrete}. We focus on the separation and obstacle avoidance by comparing six different formulations, ranging from hard barriers to slacked variants (Sections~\ref{sec:separation}
and~\ref{sec:obstacle_avoidance}).

\subsubsection{Cohesion}\label{sec:cohesion}

The \emph{active-neighbor centroid} at stage $k$ is computed as a masked weighted average of predicted neighbor positions:
\begin{equation}
  \bm{c}^{(k)} = \frac{1}{s^{(k)}}\sum_{j=1}^{K} a_j^{(k)}\,\hat{\bm{p}}_j^{(k)} \in \R^2,
  \label{eq:centroid}
\end{equation}
where the \emph{soft neighbor count}
\begin{equation}
  s^{(k)} = \max\,\left(\sum_{j=1}^{K} a_j^{(k)},\,\varepsilon\right), \qquad \varepsilon = 10^{-4},
\end{equation}
prevents division by zero when there are no neighbors in the vicinity.
The cohesion penalty is
\begin{equation}
  \ell_{\rm coh}(\bm{x}^{(k)}) =
  q_{\rm coh}\,g^{(k)}\,\|\bm{p}_i^{(k)} - \bm{c}^{(k)}\|^2,
  \label{eq:cohesion_cost}
\end{equation}
with weight $q_{\rm coh} \ge 0$, gated by the \emph{activation gate}
\begin{equation}
  g^{(k)} = \min\,\Bigl(\textstyle\sum_{j=1}^{K} a_j^{(k)},\,1\Bigr).
\end{equation}

When at least one neighbor slot is active, $g^{(k)}=1$ and the cost is enabled. In isolation, when $g^{(k)}=0$ and it vanishes. Separating $s^{(k)}$ (denominator safety) from $g^{(k)}$ (which enables the cohesion and alignment costs only when a neighbor is present) keeps the per-robot cost well posed for the SQP-RTI solve. Finally, cohesion uses the stage-matched \emph{predicted} positions $\hat{\bm{p}}_j^{(k)}$ rather than current measured ones, pulling robot $i$ toward where its neighbors are \emph{predicted to be}.

\subsubsection{Alignment}
\label{sec:alignment}

The mean heading direction ($\theta_{\rm mean}^{(k)}$) of active neighbors is estimated from their predicted velocity vectors:
\begin{align}
  \bar{\bm{v}}^{(k)} &= \sum_{j=1}^{K} a_j^{(k)}\,\hat{\bm{v}}_j^{(k)}
  = [\,\bar v_x^{(k)},\, \bar v_y^{(k)}\,]^\top, \label{eq:mean_vel}\\
  \theta_{\rm mean}^{(k)} &=
  \operatorname{atan2}\,\bigl(\bar v_y^{(k)}+\varepsilon,\;\bar v_x^{(k)}+\varepsilon\bigr). \label{eq:mean_heading}
\end{align}
The neighbor velocities are summed rather than normalized, so faster neighbors weigh more heavily. The $\varepsilon$ offsets regularize $\operatorname{atan2}$ near a zero velocity sum (fixing $\theta_{\rm mean}^{(k)}$ at $\pi/4$ only at exact cancellation, an edge case). The resulting $\theta_{\rm mean}^{(k)}$ then defines the alignment heading error, wrapped to $(-\pi,\pi]$:
\begin{equation}
  e_{\rm ali}^{(k)} = \operatorname{atan2}\,\bigl(\sin(\theta^{(k)} - \theta_{\rm mean}^{(k)}),\,\cos(\theta^{(k)} - \theta_{\rm mean}^{(k)})\bigr),
\end{equation}
and the alignment penalty is
\begin{equation}
  \ell_{\rm ali}(\bm{x}^{(k)}) = q_{\rm ali}\,g^{(k)}\,\bigl(e_{\rm ali}^{(k)}\bigr)^2,
  \label{eq:alignment_cost}
\end{equation}
with weight $q_{\rm ali} \ge 0$. As in the cohesion term, the shared gate $g^{(k)}$ disables both cohesion and alignment when robot $i$ is isolated.

\subsubsection{Inter-Robot Separation}
\label{sec:separation}

The inter-robot separation constraint builds on a pairwise safety barrier and its Dt-CBF extension. For neighbor slot $j$ at stage $k$, the barrier
\begin{equation}
  h_j^{(k)} = \bigl\|\bm{p}_i^{(k)} - \hat{\bm{p}}_j^{(k)}\bigr\|^2 - D_{\rm safe}^2
  \label{eq:barrier}
\end{equation}
is non-negative exactly when robot $i$ and neighbor $j$ are at least $D_{\rm safe}$ apart. Starting from this barrier, the same safety constraint can be encoded in several ways, each imposed \emph{hard} or \emph{slacked}. We consider six formulations that differ \emph{only} in the safety constraint, and observe the corresponding change in swarm behavior. Section~\ref{sec:comparison} then evaluates all six under identical conditions and retains the best-performing one. The last three share a common structure consisting of a slacked barrier anchor paired with a rate condition. They differ only in the rate condition, which is a one-step Dt-CBF on the body, the same condition on a forward handle, or a two-step higher-order chain.

\paragraph{(a) Hard BF}
The plain barrier imposed as a hard constraint,
\begin{equation}
  h_j^{(k)} \ge 0,
  \label{eq:v_hardbf}
\end{equation}
is the most direct encoding of the $D_{\rm safe}$ rule. It forbids any predicted
configuration closer than $D_{\rm safe}$ at every stage. It is purely reactive (it acts only once a neighbor is within range) and, as a hard constraint, allows no slack, so a robot surrounded by several neighbors at once can leave its OCP with no feasible input.

\paragraph{(b) Hard Dt-CBF}
The discrete-time CBF condition as a hard constraint,
\begin{equation}
  h_j^{(k+1)} - (1-\alpha)\,h_j^{(k)} \ge 0, \qquad \alpha\in(0,1],
  \label{eq:cbf_condition}
\end{equation}
compares the barrier $h_j^{(k)}$ with its one-step-ahead value $h_j^{(k+1)}$. The latter is formed by a forward-Euler step of the ego position and a constant-velocity step of each neighbor,
\begin{align}
  \bm{p}_i^{(k+1)} &= \bm{p}_i^{(k)} + \Delta t\, v^{(k)}[\cos\theta^{(k)},\,\sin\theta^{(k)}]^\top, \label{eq:ego_pred}\\
  \hat{\bm{p}}_j^{(k+1)} &= \hat{\bm{p}}_j^{(k)} + \Delta t\, \hat{\bm{v}}_j^{(k)}; \label{eq:nbr_pred}
\end{align}
The ego step is linear in $v^{(k)}$, which keeps the quadratic-program subproblems well conditioned. It is a first-order approximation used only inside the barrier constraint, separate from the ERK-4 discretization of the dynamics~\eqref{eq:unicycle}. The condition does not forbid $h_j<0$ but limits how fast the barrier may shrink each step, to at most a fraction $\alpha$ (a discrete class-$\mathcal{K}$ rate). The robot therefore starts slowing while still safely separated rather than braking at the boundary, giving smoother,
early avoidance. This is exactly the MPC-CBF formulation of~\citep{zeng2021mpccbf}. Being a hard constraint, it carries the same infeasibility risk as \emph{Hard BF} with dense congregations of robots.

\paragraph{(c) Slack BF}
The plain barrier mentioned in \emph{Hard BF}, softened by its slack,
\begin{equation}
  h_j^{(k)} + \xi_{j,\mathrm{bf}}^{(k)} \ge 0,
  \label{eq:v_slackbf}
\end{equation}
keeps the OCP solvable in the crowded configurations that make \emph{Hard BF} infeasible. When no input can satisfy $h_j\ge0$, the penalized slack $\xi_{j,\mathrm{bf}}$ opens just enough to admit a feasible, minimally-violating solution instead of failing.

\paragraph{(d) Slack Dt-CBF}
The slacked barrier anchor~\eqref{eq:v_slackbf} \emph{and} the discrete-time CBF condition softened by its own slack,
\begin{equation}
  h_j^{(k+1)} - (1-\alpha)\,h_j^{(k)} + \xi_{j,\mathrm{cbf}}^{(k)} \ge 0,
  \label{eq:v_slackdtcbf}
\end{equation}
imposed together on the same neighbor, each with its own slack. The rate condition combines the before-contact activation of \emph{Hard Dt-CBF} with the improved solvability of \emph{Slack BF}. The anchor guards the footprint at every stage while the CBF constraint shapes the approach rate. This is the reference point for the next two formulations: they retain this anchored structure and change only \emph{where} or \emph{how} the rate condition is evaluated, so the ensuing description only mentions those two choices.

\paragraph{(e) Look-ahead (LA) Slack Dt-CBF}
This formulation is a modification of \emph{Slack Dt-CBF} with its rate condition moved from the body onto a forward handle a fixed distance $L$ ($0<L<D_{\rm safe}$) ahead of the robot, adapting the offset-point construction of~\citet{huang2023unicyclecbf} to discrete time,
\begin{align}
  \bm{p}_{L}^{(k)} &= \bm{p}_i^{(k)} + L\,[\cos\theta^{(k)},\,\sin\theta^{(k)}]^\top, \label{eq:lookahead_pt}\\
  h_{j,L}^{(k)} &= \bigl\|\bm{p}_{L}^{(k)} - \hat{\bm{p}}_j^{(k)}\bigr\|^2 - (D_{\rm safe}-L)^2, \label{eq:lookahead}
\end{align}
with the slacked condition $h_{j,L}^{(k+1)} - (1-\alpha)\,h_{j,L}^{(k)} +
\xi_{j,\mathrm{cbf}}^{(k)} \ge 0$ imposed on the handle. A body distance barrier has relative degree two under unicycle dynamics, so the turn rate does not appear in its one-step form and a robot can only \emph{slow down} to keep clear. The handle $\bm{p}_L^{(k)}$ moves with the heading, so $\omega^{(k)}$ enters the one-step barrier directly. The robot can, therefore, also \emph{steer} around a neighbor, while the body anchor~\eqref{eq:v_slackbf} still guards the true footprint.

\paragraph{(f) Slack Dt-HOCBF}
The slacked barrier anchor~\eqref{eq:v_slackbf} together with a slacked higher-order (relative-degree-two) chain~\citep{nguyen2016exponential,xiao2019hocbf} is built \emph{natively in discrete time} from successive barrier values rather than from derivatives. With $h_j^{(k+1)}$ and $h_j^{(k+2)}$, the barrier evaluates the one-step and two-step prediction under the held input $(v^{(k)},\omega^{(k)})$,
\begin{align}
  \psi_1^{(k)} &= h_j^{(k+1)} - (1-\gamma_1)\,h_j^{(k)}, \label{eq:hocbf_psi1}\\
  \psi_2^{(k)} &= \psi_1^{(k+1)} - (1-\gamma_2)\,\psi_1^{(k)} + \xi_{j,\mathrm{cbf}}^{(k)} \ge 0, \label{eq:hocbf}
\end{align}
with decay rates $\gamma_1,\gamma_2\in(0,1]$. Expanding the chain gives the single constraint actually imposed,
\begin{equation}
  h_j^{(k+2)} - (2-\gamma_1-\gamma_2)\,h_j^{(k+1)} + (1-\gamma_1)(1-\gamma_2)\,h_j^{(k)}
  + \xi_{j,\mathrm{cbf}}^{(k)} \ge 0,
  \label{eq:dt_hocbf_expanded}
\end{equation}
where the two-step prediction advances the heading as $\theta^{(k+1)} =
\theta^{(k)} + \Delta t\,\omega^{(k)}$ and each neighbor by its constant velocity. This addresses the same relative-degree-two obstruction as~(e) but analytically rather than geometrically: $\omega^{(k)}$ is absent from the one-step barrier, yet it enters $h_j^{(k+2)}$ through the rotated heading, with sensitivity $\partial\psi_2/\partial\omega \propto \Delta t^{2} v^{(k)}$.

Across the six formulations, a safety constraint is either imposed \emph{hard} or softened by its own non-negative slack ($\xi_{j,\mathrm{bf}}^{(k)}$ for a barrier constraint, $\xi_{j,\mathrm{cbf}}^{(k)}$ for a Dt-CBF constraint), and the active slacks are penalized in the cost,
\begin{equation}
  \ell_{\rm rob\text{-}safety}^{(k)} = \sum_{j=1}^{K}\ \sum_{\xi \in \bm{\xi}_j^{(k)}} \bigl(Z^{\rm rob}\,\xi^2 + z^{\rm rob}\,\xi\bigr),
  \label{eq:rob_safety}
\end{equation}
where $\bm{\xi}_j^{(k)}$ collects the chosen formulation's inter-robot slacks for
neighbor $j$, and $Z^{\rm rob},z^{\rm rob}$ are the quadratic and linear weights. The set $\bm{\xi}_j^{(k)}$ holds both slacks for the hybrid, look-ahead, and higher-order variants~(d)--(f), a single slack for Slack BF, and none for the hard variants. Each slack is driven to zero whenever its constraint is satisfiable and absorbs only the unavoidable residual otherwise. Inactive neighbor slots are padded ($h\equiv100$) so their constraints
and slacks vanish.

\subsubsection{Obstacle Avoidance}
\label{sec:obstacle_avoidance}
The selected formulation~(a)--(f) from Section~\ref{sec:separation} is applied unchanged to static obstacles, so no new safety constraints are introduced. Each obstacle $m$ is treated as a neighbor with a fixed prediction. Three substitutions turn each inter-robot safety constraint of the chosen formulation into its obstacle counterpart. The neighbor position $\hat{\bm{p}}_j^{(k)}$ becomes the obstacle center $\bm{p}_m$ (held at zero velocity). The clearance $D_{\rm safe}$ becomes $D_{\rm safe, obs}$ (a fixed radius exceeding the obstacle radius $r_m$). The class-$\mathcal{K}$ gains take their obstacle values (defaulting to the inter-robot ones). The obstacle barrier, for instance, is
\begin{equation}
  h_m^{(k)} = \bigl\|\bm{p}_i^{(k)} - \bm{p}_m\bigr\|^2 - D_{\rm safe,obs}^2 .
  \label{eq:obs_barrier}
\end{equation}
The obstacle slacks are penalized analogously,
\begin{equation}
  \ell_{\rm obs\text{-}safety}^{(k)} = \sum_{m=1}^{M}\ \sum_{\xi \in \bm{\xi}_m^{(k)}} \bigl(Z^{\rm obs}\,\xi^2 + z^{\rm obs}\,\xi\bigr),
  \label{eq:obs_safety}
\end{equation}
with $\bm{\xi}_m^{(k)}$ the chosen formulation's obstacle slacks for obstacle $m$ and $Z^{\rm obs},z^{\rm obs}$ the quadratic and linear weights. At the terminal stage $k=N$, the one-step prediction is undefined, and all formulations revert to a plain terminal barrier.

\subsection{Look-Ahead Reference Point and Control Effort}
\label{sec:la_cost}

The look-ahead reference point $(\bm{p}^{\rm ref}, \theta^{\rm ref}, v^{\rm ref},
\omega^{\rm ref})$ produced by the behavior module of Section~\ref{sec:navigation}, whatever the mode, enters the per-robot OCP through a quadratic tracking cost. The stage-$k$ deviations of the
state from this reference point are then defined as
\begin{align}
  e_{p_x}^{(k)}    &= p_x^{(k)} - p_x^{\rm ref}, \qquad
  e_{p_y}^{(k)}     = p_y^{(k)} - p_y^{\rm ref}, \label{eq:pos_error}\\
  e_{\theta}^{(k)} &= \operatorname{atan2}\!\bigl(
  \sin\!\bigl(\theta^{(k)}-\theta^{\rm ref}\bigr),\,
  \cos\!\bigl(\theta^{(k)}-\theta^{\rm ref}\bigr)
  \bigr), \label{eq:heading_error}\\
  e_v^{(k)}        &= v^{(k)} - v^{\rm ref}, \qquad
  e_{\omega}^{(k)}  = \omega^{(k)} - \omega^{\rm ref}. \label{eq:vel_error}
\end{align}
The wrapping in~\eqref{eq:heading_error} maps the heading deviation to $(-\pi,\pi]$, avoiding discontinuities at $\pm\pi$. The deviations form a quadratic reference-tracking cost:
\begin{equation}
  \ell_{\rm la}^{(k)} =
  q_{\rm pos}\!\left({e_{p_x}^{(k)}}^{2} + {e_{p_y}^{(k)}}^{2}\right)
  + q_{\theta}\,{e_{\theta}^{(k)}}^{2}
  + q_v\,{e_v^{(k)}}^{2}
  + q_{\omega}\,{e_{\omega}^{(k)}}^{2},
  \label{eq:la_dev_cost}
\end{equation}
with weights $q_{\rm pos}, q_{\theta}, q_v, q_{\omega} > 0$. A control-effort term regularizes the inputs for smoother, more energy-efficient trajectories:
\begin{equation}
  \ell_{\rm effort}^{(k)} =
  r_u\!\left(\bigl(v^{(k)}\bigr)^{2} + \bigl(\omega^{(k)}\bigr)^{2}\right),
  \label{eq:control_cost}
\end{equation}
where $r_u > 0$.

\subsection{OCP Formulation}
\label{sec:ocp_full}

Each robot solves a finite-horizon OCP over $N$ stages. The cost assembles the look-ahead, control-effort, cohesion, and alignment terms plus CBF slack penalties into a stage and a terminal cost. This cost is minimized subject to the dynamics, the state and input box constraints, and safety constraints that enforce separation and obstacle avoidance.

\paragraph{Stage Cost}
\label{sec:stage_cost}

The stage cost at horizon stage $k \in \{0,\ldots,N-1\}$ is
\begin{equation}
  \ell_s^{(k)} = \ell_{\rm la}^{(k)}
  + \ell_{\rm effort}^{(k)}
  + \ell_{\rm coh}^{(k)}
  + \ell_{\rm ali}^{(k)}
  + \ell_{\rm rob\text{-}safety}^{(k)}
  + \ell_{\rm obs\text{-}safety}^{(k)},
  \label{eq:stage_cost}
\end{equation}
where the look-ahead, control-effort, cohesion, alignment, inter-robot slack, and obstacle slack terms are given by~\eqref{eq:la_dev_cost}, \eqref{eq:control_cost}, \eqref{eq:cohesion_cost}, \eqref{eq:alignment_cost}, \eqref{eq:rob_safety}, and~\eqref{eq:obs_safety}, respectively. Separation and obstacle avoidance no longer appear as cost terms. They are enforced through the barrier and Dt-CBF constraints of Sections~\ref{sec:separation}--\ref{sec:obstacle_avoidance}.

\paragraph{Terminal Cost}
\label{sec:terminal_cost}

The terminal cost at stage $k = N$ is
\begin{equation}
  \ell_t^{(N)} =
  q_{\rm pos}\Bigl[\bigl(p_x^{(N)} - p_x^{\rm ref}\bigr)^2
  + \bigl(p_y^{(N)} - p_y^{\rm ref}\bigr)^2\Bigr]
  + q_{\theta}\,\bigl(e_{\theta}^{(N)}\bigr)^2
  + \ell_{\rm rob\text{-}safety}^{(N)} + \ell_{\rm obs\text{-}safety}^{(N)}.
  \label{eq:terminal_cost}
\end{equation}
where the position, heading, inter-robot slack, and obstacle slack terms are given by~\eqref{eq:pos_error}, \eqref{eq:heading_error}, \eqref{eq:rob_safety}, and~\eqref{eq:obs_safety}, respectively.

\paragraph{Full OCP}
\label{sec:full_ocp}

Every robot $i$ solves the following optimization problem (with neighbor data treated as stage-wise parameters), which incorporates all the previously defined cost terms and constraints in a unified formulation:

\begingroup
\small
\begin{subequations}
  \label{eq:ocp}
  \begin{align}
    \min_{\substack{\{\bm{x}^{(k)},\bm{u}^{(k)}\}_{k=0}^{N},\\[2pt]\{\bm{\xi}_{\rm rob}^{(k)},\bm{\xi}_{\rm obs}^{(k)}\}_{k=0}^{N}}}
    \quad &
    \sum_{k=0}^{N-1} \ell_s^{(k)} + \ell_t^{(N)}
    \label{eq:ocp_obj}
    \\[0.5em]
    \text{s.t.}\quad
    & \bm{x}^{(0)} = \hat{\bm{x}}_i,
    \label{eq:ocp_init}
    \\
    & \bm{x}^{(k+1)} = F\!\bigl(\bm{x}^{(k)},\bm{u}^{(k)}\bigr), \quad k = 0,\dots,N{-}1,
    \label{eq:ocp_dyn}
    \\
    & \bm{x}_{\min} \le \bm{x}^{(k)} \le \bm{x}_{\max}, \quad k = 0,\dots,N,
    \label{eq:ocp_xbox}
    \\
    & \bm{u}_{\min} \le \bm{u}^{(k)} \le \bm{u}_{\max}, \quad k = 0,\dots,N{-}1,
    \label{eq:ocp_ubox}
    \\
    & \bm{C}_{\rm rob\text{-}safety}\bigl(\bm{x}^{(k)}, \bm{u}^{(k)};\, \{\bm{\tau}_j\}\bigr) + \bm{\xi}_{\rm rob}^{(k)} \ge \bm{0},
    \quad k = 0,\dots,N,
    \label{eq:ocp_rob}
    \\
    & \bm{C}_{\rm obs\text{-}safety}\bigl(\bm{x}^{(k)}, \bm{u}^{(k)};\, \{\bm{p}_m\}\bigr) + \bm{\xi}_{\rm obs}^{(k)} \ge \bm{0},
    \quad k = 0,\dots,N,
    \label{eq:ocp_obs}
    \\
    & \bm{\xi}_{\rm rob}^{(k)},\, \bm{\xi}_{\rm obs}^{(k)} \ge \bm{0}, \quad k = 0,\dots,N,
    \label{eq:ocp_slack}
  \end{align}
\end{subequations}
\endgroup

Here, $\bm{C}_{\rm rob\text{-}safety}$ and $\bm{C}_{\rm obs\text{-}safety}$ are the inter-robot and obstacle safety constraints of the selected formulation~(a)--(f) (Sections~\ref{sec:separation} and~\ref{sec:obstacle_avoidance}), evaluated on the neighbors' predicted trajectories $\{\bm{\tau}_j\}$ and the obstacle positions $\{\bm{p}_m\}$. The bounds $(\bm{x}_{\min},\bm{x}_{\max})$ and $(\bm{u}_{\min},\bm{u}_{\max})$ are the state and input box limits. The slack vectors $\bm{\xi}_{\rm rob}^{(k)}$ and $\bm{\xi}_{\rm obs}^{(k)}$ hold up to $2K$ and $2M$ entries for $k<N$. The predictive Dt-CBF rows require the one-step-ahead prediction, while the higher-order rows require the two-step prediction, so they drop at the terminal stage $k=N$. For the hard formulations~(a) and~(b), the slacks are fixed to $\bm{0}$.

\subsection{Simulation Workflow}
\label{sec:sim_workflow}

Each control step executes a synchronous parallel update~\cite{bertsekas2015parallel} in which all robots compute their optimal control sequences simultaneously using only neighbor information from the \emph{previous} timestep (Algorithm~\ref{alg:dmpc}). This structure prevents any robot from exploiting another robot's within-step update, matching the limited communication of a real distributed system. At the beginning of each control step, a KD-Tree~\cite{bentley1975multidimensional} is constructed over the current robot positions $\{\hat{\bm{p}}_1, \ldots, \hat{\bm{p}}_{N_b}\}$. This spatial partitioning structure enables all $K$-nearest-neighbor queries within communication range $R_{\rm comm}$ to be performed in $O(N_b \log N_b)$ total time compared to the naive pairwise distance search $O(N_b^2)$~\cite{friedman1977algorithm}. Each robot then samples the local light intensity and gradient to compute a look-ahead reference $(\bm{p}^{\rm ref}, \theta^{\rm ref}, v^{\rm ref}, \omega^{\rm ref})$ which encodes the desired position, heading, linear speed, and angular velocity according to the active swarm behavior mode. It also collects the nearby obstacle positions within range $d_{\rm obs}$ and the neighbors' predicted trajectories $\{\bm{\tau}_j\}$ buffered from the previous control step. Using these inputs, each robot solves a local OCP over horizon $N$ via a single SQP-RTI step, while updating the stage-wise parameter vector $\bm{\pi}^{(k)}$ to incorporate the look-ahead reference, neighbor trajectories, and obstacle positions as time-varying coupling parameters in~\eqref{eq:ocp}. After optimization, the robot applies the first control action $\bm{u}^{(0)}$ and publishes its updated predicted trajectory $\bm{\tau}_i$ for use by neighboring robots at time $t + \Delta t$.

\begin{algorithm}[t]
  \small
  \caption{DMPC Control Step at time $t$}
  \label{alg:dmpc}
  \begin{algorithmic}[1]
    \State Advance light field: $I(\cdot,t) \leftarrow I(\cdot,t{+}\Delta t)$
    \State \textbf{Spatial Indexing:} Construct KD-Tree from current positions
    $\{\hat{\bm{p}}_1, \ldots, \hat{\bm{p}}_{N_b}\}$

    \For{robot $i = 1,\ldots,N_b$ \textbf{in parallel}}
    \Comment{Communication Phase}
    \State Detect $K$ nearest neighbors $j \ne i$ via KD-Tree query with range
    $R_{\rm comm}$
    \State Buffer $\bm{\tau}_j = \{[\hat{\bm{p}}_j^{(k)},
    \hat{\bm{v}}_j^{(k)}]\}_{k=0}^{N}$ from neighbors'
    \emph{previous} trajectories
    \EndFor

    \For{robot $i = 1,\ldots,N_b$ \textbf{in parallel}}
    \Comment{Computation Phase}
    \State Sense light intensity $I(\hat{\bm{x}}_i)$ and gradient
    $\nabla I(\hat{\bm{x}}_i)$
    \State Compute look-ahead reference
    $(\bm{p}^{\rm ref},\, \theta^{\rm ref},\, v^{\rm ref},\, \omega^{\rm ref})$
    based on the active swarm behavior
    \State Detect the $M$ nearest obstacles within range $d_{\rm obs}$ and
    compute their center positions $\bm{p}_m$
    \State Fetch trajectories $\bm{\tau}_j$ from neighbor buffer
    \State Warm-start: $\bm{x}^{(k)} \leftarrow \hat{\bm{x}}_i$,\;
    $\bm{u}^{(k)} \leftarrow \bm{u}_{\rm nom}$ $\forall k$
    \For{$k = 0,\ldots,N$}
    \State Build and set parameter block $\bm{\pi}^{(k)}$: look-ahead reference +
    neighbor trajectory $\bm{\tau}_j$
    + obstacle positions $\bm{p}_m$
    \EndFor
    \State Execute one SQP-RTI step to obtain $\bm{u}^{(k)}$
    \State Extract first optimal control input $\bm{u}_i^* \leftarrow \bm{u}^{(0)}$ and updated
    $\bm{\tau}_i$ from solver
    \EndFor

    \State \textbf{Synchronize:} All robots publish updated trajectories $\bm{\tau}_i$
    simultaneously \Comment{Available at $t{+}\Delta t$}

    \For{robot $i = 1,\ldots,N_b$}
    \Comment{Actuation Phase}
    \State Apply $\bm{u}_i^*$
    \EndFor
  \end{algorithmic}
\end{algorithm}

\section{Simulation Results}
\label{sec:simulation}

We evaluate the proposed DMPC framework through a series of simulation studies. The studies characterize the operating density up to the geometric packing limit, compare the six inter-robot safety formulations under a common controller setup, demonstrate emergent collective behavior across the task modes, and assess real-time computational scalability with swarm size.

\subsection{Simulation Setup}
\label{sec:sim_setup}
We implemented our DMPC framework symbolically in Python using CasADi~\citep{Andersson2019} and exported the resulting OCP to acados~\citep{verschueren2022acados}, which generates and compiles highly optimized C code via its code-generation pipeline. The solver uses the acados SQP-RTI scheme, ERK-4 integration, and the interior-point QP solver HPIPM~\citep{frison2020hpipm}, and a Gauss-Newton Hessian approximation of the OCP. Simulations were performed on the compute nodes of a throughput cluster, each with two 16-core $2.90\,\mathrm{GHz}$ Intel Xeon Gold 6326 (Ice Lake) processors, i.e.\ 32 physical cores, and $250\,\mathrm{GB}$ of RAM. Every run was pinned to a single core, so the per-robot solve times reported in Section~\ref{sec:performance} are single-core numbers.

The physics environment was simulated using a custom 2D engine within a fixed
$0.8\,\mathrm{m}\times 0.8\,\mathrm{m}$ workspace (horizon $N=20$, control step $\Delta t=0.1\,\mathrm{s}$, up to $K=5$ neighbors). The NMPC and Reynolds penalty weights were tuned via Bayesian optimization with Optuna~\citep{optuna_2019} to maximize swarm coherence while prioritizing safety and solver stability. The lower speed bound $v_{\min}=0.01$\,m/s (Table~\ref{tab:params}) keeps every robot moving instead of letting it stop. The environmental Gaussian light source rotated along a circular trajectory in the phototaxis experiment and remained stationary in the phototactic orbital experiment. All robots were initialized at uniform random positions in a central sub-region covering 90\% of the arena $\mathcal{A}$ with a minimum initial separation of $D_{\rm safe}$ and ran for 9000 control steps ($T_{\rm sim} = 900\,\mathrm{s}$), long enough for the swarm to reach and hold a steady state rather than being measured mid-transient.

\paragraph{Metrics and protocol}
All reported quantities are means over $10$ random seeds with $95\%$ confidence intervals. Steady-state quantities are averaged over the last $3000$ of the $9000$ steps (the final $300\,\mathrm{s}$). A robot-robot (or robot-obstacle) \emph{contact} is a distinct overlap of the elliptical footprints, evaluated at the MPC-commanded positions and counted once per overlap episode. All runs include a lightweight reactive collision-response layer (a per-robot heading bias plus a few positional-separation sweeps) matching the deployed system. A contact is detected as soon as the footprints overlap, before the layer reacts. The layer then pushes the overlapping pair apart and biases their headings away from each other, so overlaps stay shallow and brief. Solver feasibility is the fraction of per-robot solves returning a successful acados status. The six formulations share an identical stack, weights, horizon, arena, seed set, and safe distance $D_{\rm safe}$. Only the inter-robot safety and obstacle-avoidance constraints differ. For the look-ahead formulation, the pair $(\alpha, L)$ was selected by a grid screening followed by Bayesian optimization, and for the anchored Slack Dt-CBF, the decay rate $\alpha$ was selected by a direct two-point comparison, each re-verified over the full ten-seed protocol. Formulations~(d) and~(e) therefore share $\alpha$, which is what makes the look-ahead ablation below a single-variable comparison. The remaining formulations run at the gains inherited from their source implementations rather than at individually optimized ones; a single shared gain would not be meaningful across formulations that differ in structure, and we report this asymmetry because it bounds how the cross-formulation ranking should be read. The complete set of cost weights, CBF gains, slack weights, and environment constants is listed in Appendix~\ref{app:supplementary_data},Table~\ref{tab:params}.

\subsection{Capacity and Operating Density}
\label{sec:capacity}
We now characterize how densely the swarm can pack in the fixed arena. This section establishes the geometric capacity and selects the operating point used in the later analysis.

\paragraph{Geometric capacity}
The swarm size is limited by the safety buffer, not the physical footprint. The elliptical footprint is small, only $\approx 11\%$ of the arena at $N_b=50$, against $\approx 30\%$ when the buffers are also considered. At $D_{\rm safe}=7\,\mathrm{cm}$, the buffers cap the swarm size at an upper limit of $N_{\max}=137$ non-overlapping safety disks of radius $D_{\rm safe}/2$ (Table~\ref{tab:swarm_density_and_occupancy}). This packing results in $\approx 82\%$ area coverage, just below the ideal hexagonal $90.7\%$ because of boundary and finite-size effects. The ceiling depends on the safety distance, ranging from $N_{\rm max}=188$ at $D_{\rm safe}=6\,\mathrm{cm}$ to $N_{\rm max}=68$ at $D_{\rm safe}=10\,\mathrm{cm}$. Across these distances, the buffer occupancy at the limit stays near $82$--$83\%$. We therefore use buffer occupancy as a density measure comparable across safety distances.

\paragraph{Operating point}
We select the operating point with a closed-loop DMPC sweep over $D_{\rm safe}$ and movement headroom $\Delta N$. The sweep scores occupancy, collisions, feasibility, solve time, and neighbor-buffer utilization. Sweeping over $D_{\rm safe}\in\{7,8\}\,\mathrm{cm}$ over the four barrier and CBF formulations available at the time, it finds both viable, so we set $D_{\rm safe}=D_{\rm safe, obs}=7\,\mathrm{cm}$, the tighter of the two. The working maximum $N_b=100$ then follows from the geometric headroom criterion $N_{\rm op}=N_{\max}-\Delta N$ with $\Delta N=37$ ($\approx 60\%$ buffer occupancy), not from the sweep score. Beyond $N_{\max}$ the safety buffers can no longer be packed without forced overlap, and the swarm stays feasible only through the slacked barrier.

\begin{table}[pos=htbp]
  \centering
  \small
  \caption{Swarm size, physical robot footprint, and safety buffer occupancy in a $0.8\,\mathrm{m} \times 0.8\,\mathrm{m}$ ($0.64\,\mathrm{m^2}$) arena. The physical footprint uses $A_{\rm bot} \approx 14.14\,\mathrm{cm}^2$ and the safety buffer area is $A_{\rm buffer}\approx 38.48\,\mathrm{cm}^2$ ($D_{\rm safe} = 7\,\mathrm{cm}$). The green row ($N_b = 50$) marks the representative swarm size used for the behavior figures. The yellow row ($N_b = 100$) marks the largest swarm size that leaves some headroom to move. The red row marks the geometric non-overlapping packing limit ($N_{\max} = 137$).}
  \label{tab:swarm_density_and_occupancy}
  \begin{tabular}{ccccc}
    \toprule
    & \multicolumn{2}{c}{\textbf{Physical Robot Footprint}} & \multicolumn{2}{c}{\textbf{Safety Buffer}} \\
    \cmidrule(lr){2-3} \cmidrule(lr){4-5}
    \textbf{Swarm Size} & \textbf{Total Area} & \textbf{Occupancy} & \textbf{Total Area} & \textbf{Occupancy} \\
    $N_b$ & ($\mathrm{cm}^2$) & (\%) & ($\mathrm{cm}^2$) & (\%) \\
    \midrule
    1   & 14.14   & 0.22  & 38.48   & 0.60  \\
    10  & 141.40  & 2.21  & 384.85  & 6.01  \\
    25  & 353.50  & 5.52  & 962.11  & 15.03 \\
    \rowcolor{green!10}
    50  & 707.00  & 11.05 & 1924.23 & 30.07 \\
    75  & 1060.50 & 16.57 & 2886.34 & 45.10 \\
    80  & 1131.20 & 17.68 & 3078.76 & 48.11 \\
    \rowcolor{yellow!10}
    \textbf{100} & \textbf{1414.00} & \textbf{22.09} & \textbf{3848.45} & \textbf{60.13} \\
    120 & 1696.80 & 26.51 & 4618.14 & 72.16 \\
    \rowcolor{red!20}
    \textbf{137} & \textbf{1937.18} & \textbf{30.27} & \textbf{5272.38} & \textbf{82.38} \\
    \bottomrule
  \end{tabular}
\end{table}

\subsection{Safety Formulation Comparison}
\label{sec:comparison}
\begin{figure}[pos=htbp]
  \centering
  \includegraphics[width=\linewidth]{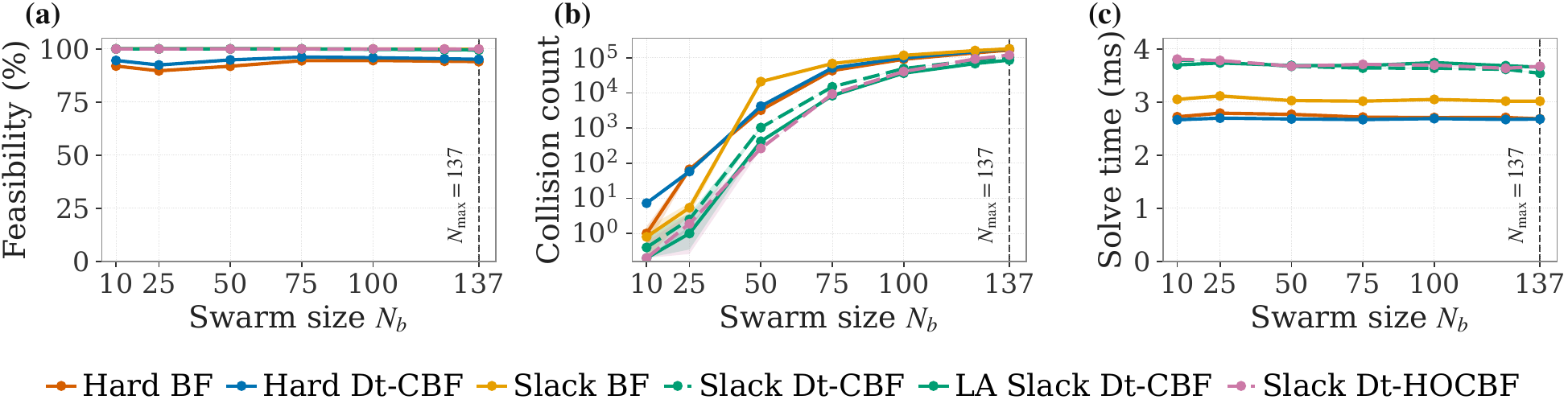}
  \caption{Safety-formulation comparison across the six variants (phototaxis, free space,
  10-seed mean, $95\%$ CI bands). (a) Solver feasibility, (b) inter-robot collisions
  (distinct contacts over 9000 steps, log scale), and (c) mean per-robot QP solve time,
  versus swarm size $N_b$. The dashed vertical line marks the geometric packing limit
  $N_{\max}=137$.}
  \label{fig:comparison}
\end{figure}

The inter-robot and obstacle safety constraints can be formulated in several ways. To map this design space, we compare the six formulations~(a)--(f) of Sections~\ref{sec:separation} and~\ref{sec:obstacle_avoidance} for the phototaxis mode in the absence of obstacles. The slacked variants carry the feasibility-recovery idea of~\citet{zeng2021feasibility} over to our distributed swarm. Under the identical setup of Section~\ref{sec:sim_setup} (same stack, solver, and closed-loop simulator), the six formulations differ only in how they enforce inter-robot and obstacle safety, so the comparison isolates that single design choice. Each configuration is evaluated at $N_b \in \{10,25,50,75,100,125,137\}$ over 10 seeds, with $N_{\max}=137$ marked by the dashed vertical line in the figures. Fig.~\ref{fig:comparison} reports the solver feasibility, the collisions, and the mean QP solve time versus swarm size $N_b$, and Table~\ref{tab:comparison} summarizes the low-density operating point $N_b = 25$ alongside a high-density operating point $N_b = 100$.

\begin{table}[pos=htbp]
  \centering
  \small
  \caption{Safety-formulation comparison (phototaxis, free space, 10-seed mean $\pm$ $95\%$ CI). Collisions are cumulative counts of distinct robot-robot contacts over 9000 steps, reported at a low density ($N_b=25$) and at a high-density operating point $N_b=100$. Feasibility and solve time are at $N_b=25$. Bold marks the best value in each column.}
  \label{tab:comparison}
  \begin{tabular}{lcccc}
    \toprule
    \textbf{Variant} & \textbf{Feas. (\%)} & \textbf{Solve (ms)} & \textbf{Coll. $N_b{=}25$} & \textbf{Coll. $N_b{=}100$} \\
    \midrule
    Hard BF                                & $89.70 \pm 0.03$ & $2.79 \pm 0.01$ & $67 \pm 8$ & $90608 \pm 701$ \\
    Hard Dt-CBF~\citep{zeng2021mpccbf}     & $92.44 \pm 0.05$ & $\mathbf{2.70 \pm 0.01}$ & $58 \pm 7$ & $99704 \pm 869$ \\
    Slack BF                               & $100.00 \pm 0.00$ & $3.11 \pm 0.04$ & $5 \pm 2$ & $116979 \pm 417$ \\
    Slack Dt-CBF                           & $100.00 \pm 0.00$ & $3.77 \pm 0.01$ & $2 \pm 1$ & $49251 \pm 614$ \\
    \textbf{LA Slack Dt-CBF}               & $100.00 \pm 0.00$ & $3.74 \pm 0.02$ & $\mathbf{1 \pm 1}$ & $\mathbf{36505 \pm 493}$ \\
    Slack Dt-HOCBF                         & $100.00 \pm 0.00$ & $3.78 \pm 0.01$ & $2 \pm 2$ & $40628 \pm 807$ \\
    \bottomrule
  \end{tabular}
\end{table}

\paragraph{Feasibility}
Solver feasibility separates the hard and slacked formulations most sharply (Fig.~\ref{fig:comparison}a). The two hard formulations hold only about $90\%$ to $92\%$ feasibility at $N_b = 25$. When a solver is infeasible (roughly $8\%$ to $10\%$ of the per-robot OCPs), the robot falls back to a reactive potential-field controller (goal attraction with inter-robot, obstacle, and wall repulsion). This matches the recursive-infeasibility limitation that \citet{zeng2021mpccbf} identify for the hard discrete-time CBF. Softening the barrier with slack variables removes the problem. All four slacked formulations (Slack BF, Slack Dt-CBF, LA Slack Dt-CBF, and Slack Dt-HOCBF) hold at least $99.4\%$ feasibility across the entire density range, and $100\%$ up to $N_b=25$; the minimum over all densities and all four is $99.47\%$, at the packing limit. The small remaining infeasibility comes from the hard arena-boundary constraint, not the safety constraint. At high density, robots pressed against the arena walls can briefly make it infeasible. Feasibility alone, however, does not separate these four. As the collision results show next, a formulation can be feasible everywhere yet still fail to keep robots apart.

\paragraph{Collisions}
The formulations diverge in collision counts, not feasibility
(Fig.~\ref{fig:comparison}b), and they separate by \emph{structure} rather than by
barrier order. At $N_b=100$ the three formulations that pair a barrier anchor with a rate
condition~(d)--(f) average $4.2\times10^{4}$ contacts, against $1.02\times10^{5}$ for the
three without an anchor~(a)--(c), a factor of $2.4$. The two groups do not overlap at any
density in the sweep. Carrying an anchor alongside the rate condition therefore matters
more than which rate condition is used.

\begin{figure}[pos=htbp]
  \centering
  \includegraphics[width=\linewidth]{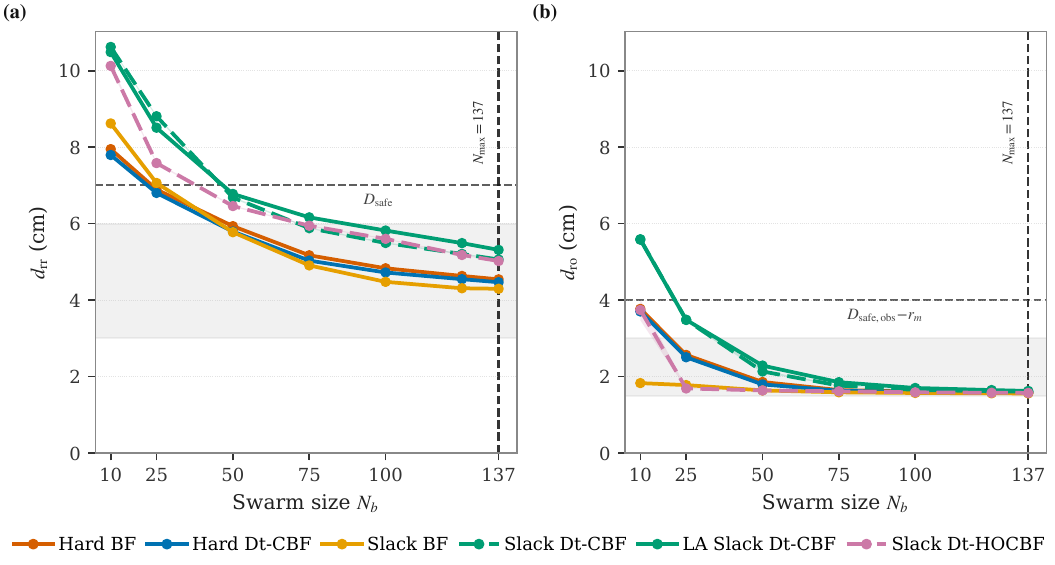}
  \caption{Safety margins of the six safety formulations versus swarm size (phototaxis with two static obstacles, deployed closed loop with the reactive collision-response layer active,
  10-seed mean, $95\%$ CI bands). (a) Steady-state mean nearest-neighbor distance $d_{\mathrm{rr}}$, center to center, dashed horizontal line at $D_{\rm safe}=7$\, cm. (b) Mean robot-obstacle clearance $d_{\mathrm{ro}}$, robot center to obstacle surface, dashed horizontal line at $D_{\rm safe, obs}-r_m=4$\, cm, since the obstacle barrier bounds the center-to-center distance. The gray band spans the footprint-contact range, which varies with orientation for elliptical robots ($2b$--$2a$ between two robots, $b$--$a$ from a robot center to an obstacle surface). The vertical dashed line marks the geometric packing
  limit.}
  \label{fig:distances}
\end{figure}

Two further results follow within the anchored group. First, the look-ahead handle alone
reduces contacts by a factor of $2.5$ at low and moderate density, falling to $1.2$ at the
packing limit, where it takes them from $1.05\times10^{5}$ to $8.6\times10^{4}$.
Formulations~(d) and~(e) are identical apart from the handle, at the same decay rate
$\alpha=0.9$, with disjoint confidence intervals at every density $N_b\ge50$ and no
measurable solve-time cost. The reduction is therefore due to the handle, which gives the
angular velocity direct authority over the constraint so a robot can steer around a
neighbor rather than only brake. Second, the higher-order chain helps at moderate density
and stops helping at the packing limit. The Slack Dt-HOCBF is second best in small swarms,
but by $N_b=137$ it falls behind the LA Slack Dt-CBF, $1.19\times10^{5}$ against
$8.6\times10^{4}$. It does not fail by becoming infeasible, holding $100\%$ feasibility
across the whole range against $99.5\%$ for~(d) and~(e). What it gives up is margin, and
not just at the packing limit. The chain lets its robots run closer than the look-ahead
variant at every density (Fig.~\ref{fig:distances}a), $5.02$ against
$5.31\,\mathrm{cm}$ at the packing limit. Its two-step condition allows the barrier to
shrink further before the robot has to act. That costs nothing when the arena has
unoccupied space. As the available area shrinks, a smaller gap leads to contact. On this
task, the deployed LA Slack Dt-CBF is best or tied-best at every density except $N_b=50$,
where it is second behind the Slack Dt-HOCBF with disjoint confidence intervals. Elsewhere,
the look-ahead advantage over the chain widens as the arena fills, from $1.11\times$ at
$N_b=100$ to $1.39\times$ at $N_b=137$.

\paragraph{Safety margins}
The formulations also differ in the safety \emph{margin} they sustain (Fig.~\ref{fig:distances}). These runs use the deployed closed loop with the reactive layer active (Section~\ref{sec:sim_setup}), and are measured on post-steering positions, so they never show interpenetration. Each value is a mean over robots and steps of the nearest-neighbor distance, not a per-step minimum. The margins separate along the same structural line as the collision counts. On the inter-robot axis, the three anchored formulations~(d)--(f) hold $10.1$--$10.6\,\mathrm{cm}$ at $N_b=10$ and $5.0$--$5.3\,\mathrm{cm}$ at the packing limit. The three without an anchor~(a)--(c) fall from $7.8$--$8.6\,\mathrm{cm}$ to $4.3$--$4.5\,\mathrm{cm}$. Comparing group means, the anchor sustains roughly $0.7\,\mathrm{cm}$ of extra clearance at maximum crowding. The deployed LA Slack Dt-CBF is widest of all at high density, with $5.82\,\mathrm{cm}$ at
$N_b=100$ and $5.31\,\mathrm{cm}$ at $N_{\max}=137$. At $N_b=25$, only the anchored formulations still hold a clear margin above $D_{\rm safe}$, at $7.6$--$8.8\,\mathrm{cm}$. The unanchored group sits at $6.8$--$7.1\,\mathrm{cm}$, at or just either side of it.
Beyond that density, every formulation operates inside the nominal buffer, already $6.5$--$6.8$ against $5.8$--$5.9\,\mathrm{cm}$ at $N_b=50$. The task drives this, not the choice of formulation. Phototaxis gathers the swarm at the light peak, so the robots there
sit far closer than the arena average of $30\%$ buffer coverage at $N_b=50$ (Table~\ref{tab:swarm_density_and_occupancy}) would suggest, and neighboring buffers have to overlap. What the anchor leads to is a consistently larger share of the space that remains, and from $N_b=75$ upward the formulation that holds the widest spacing also records the fewest contacts.

On the robot-obstacle axis, the ordering is similar, but the effect concentrates at the low-density end. At $N_b=10$ the Slack Dt-CBF and the LA Slack Dt-CBF~(d),(e) keep about $5.6\,\mathrm{cm}$ of clearance each, against $1.8$--$3.8\,\mathrm{cm}$ for the remaining four, the Slack Dt-HOCBF included. From $N_b=100$ and up, every formulation converges to $1.6$--$1.7\,\mathrm{cm}$. At that density, robots are pressed against obstacles by their neighbors rather than by their own control choices, so the safety formulation can no longer set the obstacle clearance. This is the margin-side counterpart of the collision result that obstacles stop mattering once inter-robot congestion dominates (Section~\ref{sec:density_comparison}).

\begin{figure}[pos=htbp]
  \centering
  \includegraphics[width=\linewidth]{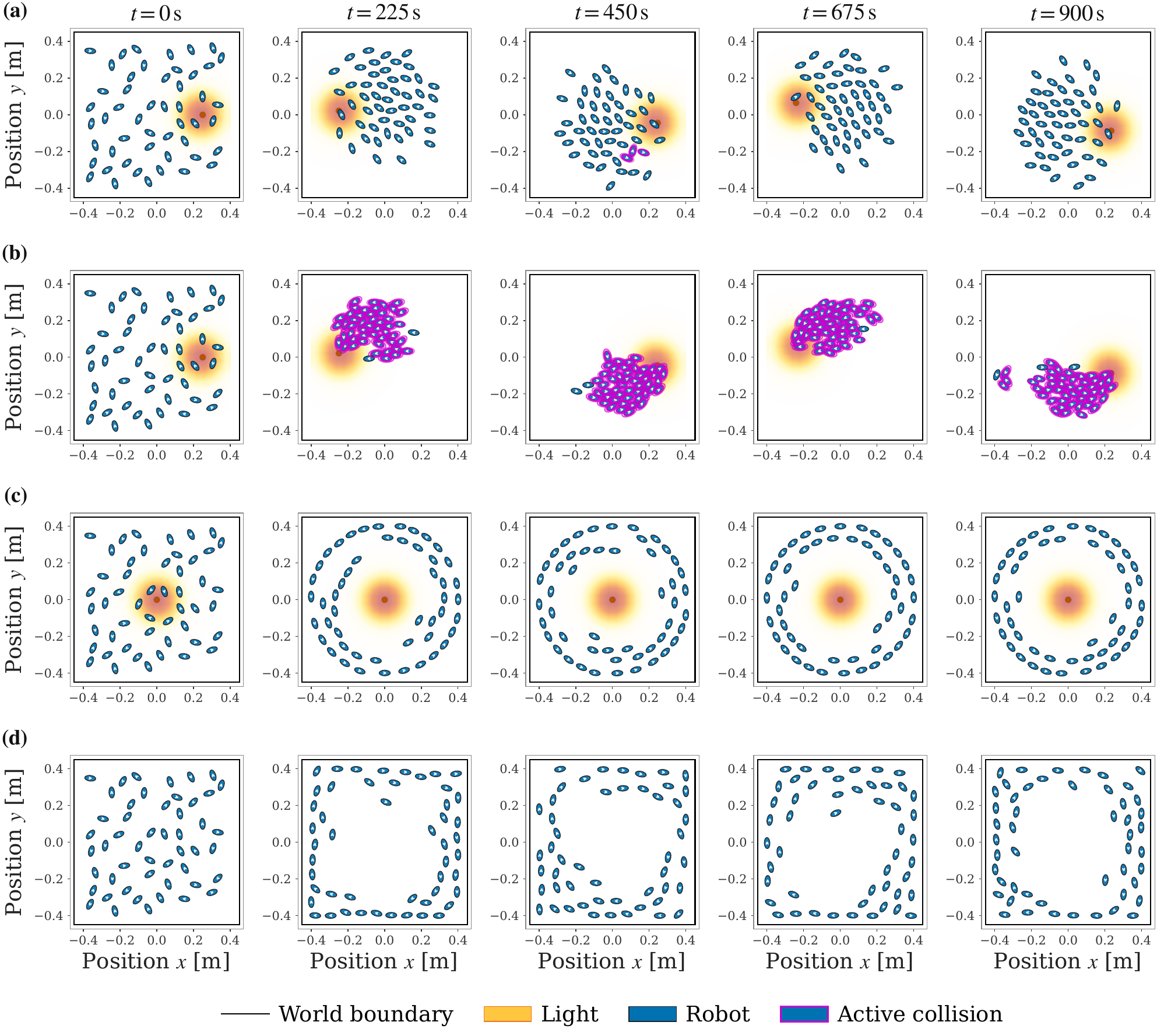}
  \caption{Collective behavior in obstacle-free space with the LA Slack Dt-CBF formulation, $N_b=50$. (a) Phototaxis, where the swarm stays cohesive and heading-aligned while tracking the moving light source. (b) Phototaxis with the coupling costs and avoidance constraints removed, where the robots pile up at the light peak and collide. (c) Phototactic orbiting, where the swarm co-rotates along the light's intensity level set. (d) Cohesion without a stimulus, where the Reynolds costs alone keep the group cohesive. Each column corresponds to a certain time $t$, and the orange field is the light source. Supplementary video~1 shows these four behaviors in the obstacle-free arena.}
  \label{fig:behaviors}
\end{figure}

\paragraph{Computational cost}
The LA Slack Dt-CBF costs slightly more to solve because of its added look-ahead constraints. Its per-robot solve time is about $3.7\,\mathrm{ms}$ at $N_b = 25$, roughly $1.0\,\mathrm{ms}$ above the cheapest formulations (Fig.~\ref{fig:comparison}c). Each
robot optimizes against a fixed number of neighbors, so this per-robot cost barely changes with swarm size and stays within $3.66$--$3.74\,\mathrm{ms}$ across the sweep. It is therefore well inside the $100\,\mathrm{ms}$ budget of a $10\,\mathrm{Hz}$ loop at every
density. Section~\ref{sec:performance} reports the distribution and separates this per-robot cost from the per-step wall time of the single-process simulation, which does grow with $N_b$. The extra cost is small, so on phototaxis, the deployment target, the LA Slack Dt-CBF gives the best trade-off across feasibility, collisions, and safety margin. Therefore, we adopt it as the deployed formulation for the collective behavior simulations. 

\subsection{Emergence of Collective Behavior}\label{sec:collective_behavior}
We characterize the emergent behavior of the swarm under the DMPC framework. The LA Slack Dt-CBF formulation governs both inter-robot separation and obstacle avoidance. We study three tasks, namely coordinated phototaxis, phototactic orbiting, and cohesion without a stimulus. Fig.~\ref{fig:behaviors} shows representative snapshots at $N_b=50$. The snapshots show how the local coupling terms regulate spatial density and heading agreement while each agent pursues its objective, without global communication or a central coordinator. With two and with twelve static obstacles in the arena, every behavior still holds its form while the swarm moves around the obstacles (Figs.~\ref{fig:behaviors_2obs} and~\ref{fig:behaviors_12obs} in Appendix~\ref{app:behaviors_obstacles}). 

We assess coordination among the robots using two distinct metrics. The first is the polar order parameter calculated locally over only the set of active neighbors,
\begin{equation}
  \Phi_{\rm loc} = \frac{1}{N_b} \sum_{l=1}^{N_b}\frac{1}{K_{l}}
  \Bigl\lVert\sum_{j \in \mathfrak{N}_{l}} e^{i\theta_j}\Bigr\rVert \in [0,1],
  \label{eq:local_polar_order}
\end{equation}
where $\mathfrak{N}_{l}$ is the set of active neighbors of each robot $l$ inclusive of the robot $l$, and
$K_l = |\mathfrak{N}_l|$ is the number of active neighbors for $l$. When all the robots in every neighborhood are in near-perfect
alignment, $\Phi_{\rm loc} \approx 1$, whereas in a disordered state with random robot
orientations, $\Phi_{\rm loc} \approx 0$. This metric is widely used in biophysics to
quantify order in a moving flock or swarm~\cite{couzin2002collective,
vicsek2012collective}. The second metric is the nearest-neighbor distance averaged over all robots,
\begin{equation}
  \mathrm{NND} = \frac{1}{N_b}\sum_{l=1}^{N_b} \lVert\bm{x}_l - \bm{x}_{k}\rVert,
  \label{eq:nnd}
\end{equation}
where $k$ is the nearest neighbor of robot $i$. This metric finds pervasive
use in quantifying internal group structure in natural and artificial
flocks~\cite{partridge1980sensory, berlinger2021implicit}. While the NND varies with
animal species, it unequivocally reaches a stable value at long time scales. The top and
bottom rows of Fig.~\ref{fig:order_and_nnd} elucidate the variation in the local polar
order and long-time NND for different swarm sizes and different obstacle numbers, under
distinct modes of operation.

\subsubsection{Phototaxis Behavior}
We first examine the full DMPC formulation with all Reynolds steering rules (Section~\ref{sec:reynolds}) active, demonstrating gradient-following behavior, followed by isolating the contribution of the social coupling by removing these terms entirely.

\paragraph{With Reynolds Flocking Rules -- Collective Behavior} \label{sec:with_reynolds}
In phototaxis mode, each robot ascends the light gradient while the cohesion and alignment costs, together with the CBF separation constraints, produce collective motion. The CBF
constrains the closing velocity toward each neighbor's \emph{predicted} position, so avoidance starts early, without the over-braking or deadlocks of purely reactive
methods~\citep{chandra2025deadlock}. The swarm stays compact and heading-aligned as it tracks the rotating source (Fig.~\ref{fig:behaviors}a). The local polar order reduces with increasing swarm size, as radial phototaxis uses a convergent field that brings agents together from all directions, producing head-on encounters that locally cancel heading agreement (Fig.~\ref{fig:order_and_nnd}a). A cluttered environment further worsen the alignment by introducing additional collision avoidance, which de-aligns the local heading. The NND (Fig.~\ref{fig:order_and_nnd}d), on the other hand, decreases as the space for movement in the domain decreases (increasing swarm size), partly also attributable to the nature of the phototaxis mode, which attempts to converge the robots to a small location. Beyond $N_b=50$, the NND falls below the safe buffer distance of $D_{\rm safe} = 7\, \mathrm{cm}$, indicating some amount of buffer overlap. As the swarm size approaches the geometric packing limit, the phototaxis gradient forces the robots into oblique, blocking contacts, causing NND to fall as low as $5\,\mathrm{cm}$.

\paragraph{Without Reynolds Flocking Rules -- Swarm Collapse}
\label{sec:no_reynolds}
In the absence of the cohesion and alignment costs and the separation constraints, the collective behavior disappears. Each
robot climbs the gradient on its own, so the swarm arrives at the light peak without spacing or heading agreement (Fig.~\ref{fig:behaviors}b). The swarm effectively collapses onto a small area in the domain, clearly attributing the occurrence of flocking to the Reynolds steering costs.

\subsubsection{Phototactic Orbital Behavior}
In the phototactic orbital mode, each robot follows the tangent of the sensed intensity level set through the gradient-rotation reference~\eqref{eq:orbit_ref}, orbiting the source without global localization. The CBF constraints~\eqref{eq:lookahead} prevent collisions as robots compete for the same orbit while the cohesion and alignment costs keep the swarm compact and co-rotating. From arbitrary initial positions, it converges to
a stable orbital formation (Fig.~\ref{fig:behaviors}c). Although the phototactic orbital mode is susceptible to crowding similar to the phototaxis mode, the decay of the local polar order is more muted in the former (Fig.~\ref{fig:order_and_nnd}b). This is a direct consequence of the co-location of local neighbors along concentric orbital paths, leading to better heading agreement, even for large swarms. Similarly, the NND also settles slightly below the safe buffer distance even for $N_b=137$, as the locally tangential motion along the orbital causes the robots to align along their major axes, allowing for sufficient separation among them (Fig.~\ref{fig:order_and_nnd}e). Both metrics are only mildly affected by the addition of obstacles to the system.

\subsubsection{Flocking in the Absence of Stimuli}
\label{sec:cohesion_without_stimulus}
In this mode, the light field is disabled, providing the robots no global stimulus. Hence, each robot holds its heading with a small random turn each step. The swarm drifts outward and settles into a band along the walls, with an empty interior (Fig.~\ref{fig:behaviors}d). Once a robot reaches a wall, it stays there until its heading drifts away, so the walls
fill faster than they empty. Cohesion and alignment order the band, so the swarm circulates the arena similar to an edge current~\citep{beppu2021edge}. 

\begin{figure*}[pos=htbp]
  \centering
  \includegraphics[width=\linewidth]{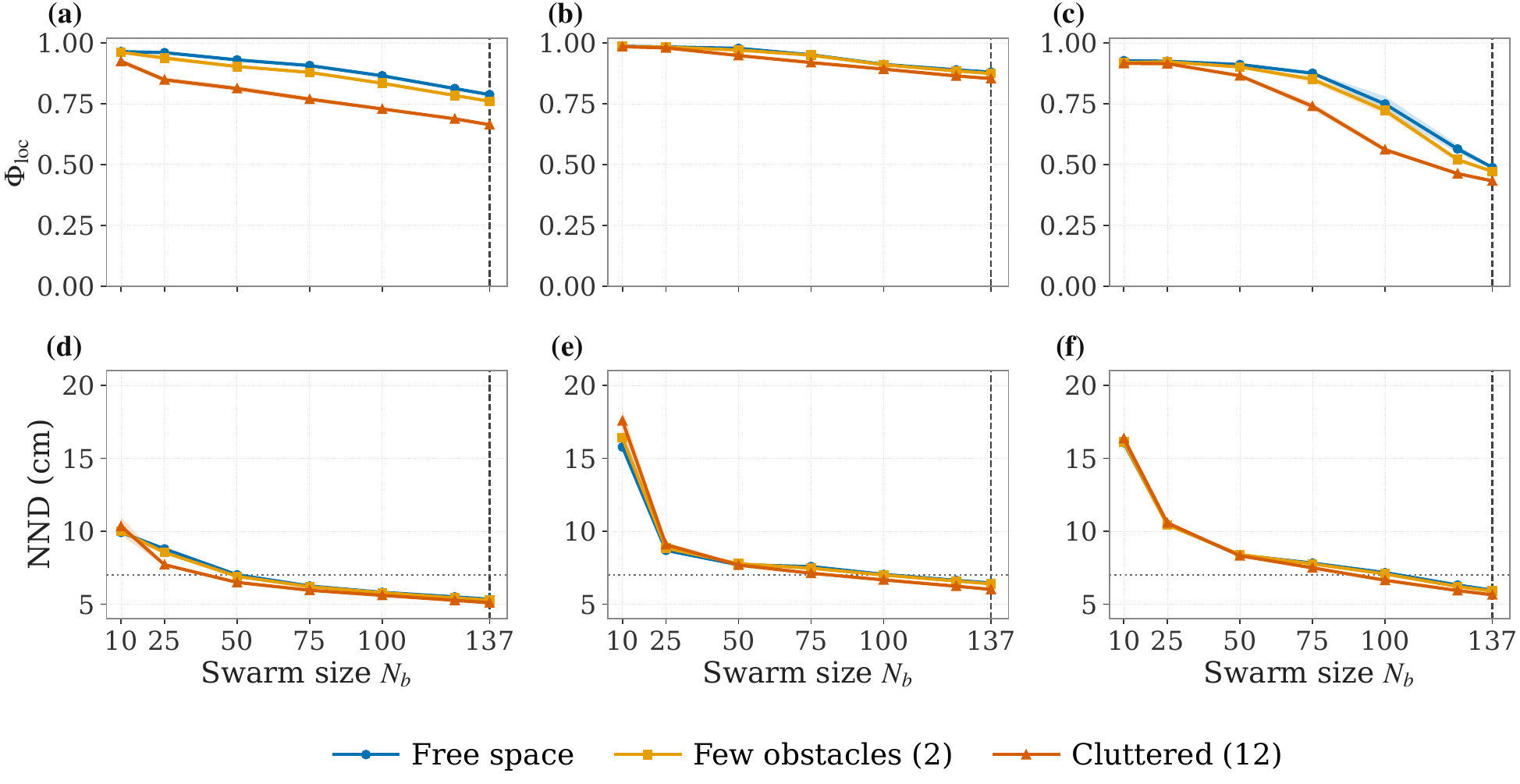}
  \caption{Local polar order $\Phi_{\rm loc}$ (top row) and average nearest-neighbor distance (NND; bottom row) versus swarm size for the deployed LA Slack Dt-CBF, one column per collective behavior (10-seed mean with $95\%$ CI bands, steady state over the last $300\,\mathrm{s}$). The columns are phototaxis (a, d), phototactic orbiting (b, e), and cohesion without a stimulus (c, f). Each panel compares the obstacle-free arena, the few-obstacle arena ($2$ obstacles), and the cluttered arena ($12$ obstacles, positions randomized per seed). The dotted horizontal line marks the buffer distance of $D_{\rm safe}=7\,\mathrm{cm}$, while the vertical dashed line marks the geometric packing limit $N_{\max}=137$.}
  \label{fig:order_and_nnd}
\end{figure*}

The flock remains relatively well-aligned at low swarm sizes, with heading disagreements as the domain starts getting crowded with robots and obstacles due to frequent steering for collision avoidance (Fig.~\ref{fig:order_and_nnd}c). This is an expected outcome, showing that the alignment and cohesion costs work as intended. The band structure resembles the phototactic orbital, despite the difference in shape of the orbit. This is evident from the NND (Fig.~\ref{fig:order_and_nnd}f), which follows a similar trend as in the phototactic orbital mode.

\subsection{Collision Behavior Across Density and Tasks}
\label{sec:density_comparison}
The collision load of the deployed LA Slack Dt-CBF depends on both swarm size and task.
We measure both here, across three obstacle environments.

\begin{figure}[pos=htbp]
  \centering
  \includegraphics[width=\linewidth]{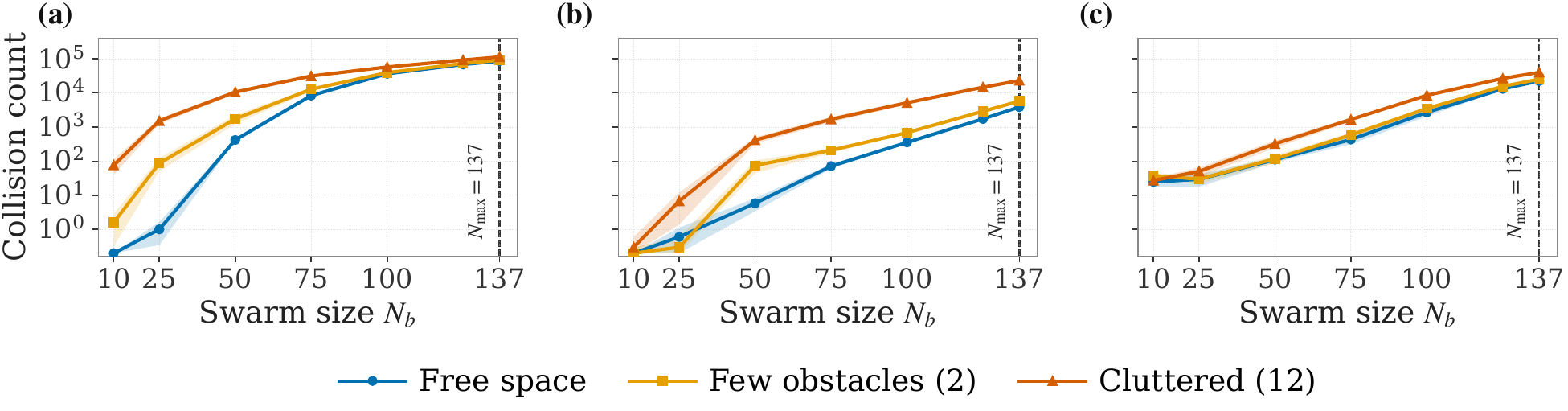}
  \caption{Distinct contacts versus swarm size for the deployed LA Slack Dt-CBF. (a) Phototaxis, (b) phototactic orbiting, and (c) cohesion without a stimulus. Each panel compares the obstacle-free arena, the few-obstacle arena ($2$), and the cluttered arena ($12$, positions randomized per seed). The vertical axis is log-scaled. The dashed vertical line marks the packing limit $N_{\max}=137$. The band around each curve is the $95\%$ confidence interval over 10 seeds.}
  \label{fig:collisions}
\end{figure}

\paragraph{Collisions versus swarm size}
We sweep $N_b\in\{10,25,50,75,100,125,137\}$ across three environments, namely
obstacle-free, few-obstacle ($2$), and cluttered ($12$, positions randomized per seed).
We count distinct robot-robot contacts over the $9000$-step run (Fig.~\ref{fig:collisions}). Contacts stay negligible in small swarms. Free space averages
$1.0$ contact at $N_b=25$, with the nearest neighbor $8.8\,\mathrm{cm}$ away on average,
well above $D_{\rm safe}$. Contacts then climb steeply, from $4.2\times10^{2}$ at
$N_b=50$ to $8.6\times10^{4}$ at the packing limit, where the mean nearest-neighbor
separation has fallen to $5.3\,\mathrm{cm}$. The growth is strongly linear. Contacts
rise by a factor of $87$ between $N_b=50$ and $N_b=100$ while the robot count only
doubles. Near the packing limit the full $D_{\rm safe}$ margin cannot be held for every
pair, so the slacked CBF admits transient contacts while the solver stays feasible
(Section~\ref{sec:comparison}).

Obstacles dominate in small swarms and hardly matter in large ones
(Fig.~\ref{fig:collisions}). At $N_b=25$ the cluttered arena produces three orders of
magnitude more contacts than free space. By $N_b=100$ that gap has closed to a factor of
$1.6$, and by $N_b=125$ to $1.3$. The crossover falls between $N_b=50$ and $N_b=75$. The environment therefore dominates in small swarms, while in large swarms collisions come
from the packing density.

\paragraph{Effectiveness of avoidance is set by task geometry}
Considering an identical controller, the number of contacts spans two orders of magnitude across the three behaviors, a factor of $103$ between the best and the worst at $N_b=100$ (Fig.~\ref{fig:collisions}). Phototaxis concentrates the swarm at the peak intensity of the light source and has the highest number of contacts at $3.7\times10^{4}$. The orbital mode spreads robots along an intensity level set and has far fewer contacts at $3.5\times10^{2}$. Cohesion without a stimulus falls in between at $2.7\times10^{3}$. However, this order of effectiveness is not preserved across different swarm sizes. The stimulus-free mode is the worst in small swarm, with $25$ contacts at $N_b=10$, whereas the other modes record none. The cause is how robots meet, not how crowded they are. At $N_b=10$ the stimulus-free swarm is the most spread out of the three, with a mean nearest-neighbor distance of $16.1\,\mathrm{cm}$ against $9.9\,\mathrm{cm}$ for phototaxis. The difference is in their headings. A shared gradient gives phototaxis and orbiting a common direction, so two robots that come close are already traveling the same way. Without a stimulus each robot follows its own random heading, and only the local coupling keeps neighbors together. In large swarms the same absence of a shared target protects it, since the robots never concentrate. At every larger swarm size orbiting stays safest and phototaxis worst, except in the cluttered arena where the stimulus-free mode moves ahead of orbiting at $N_b=50$ and $N_b=75$. Obstacles raise contacts in every mode, but the orbital mode pays the largest penalty, $14.5\times$ at $N_b=100$ against $1.6\times$ for phototaxis. Obstacles sit directly on the orbit and force detours. The safety formulation decides how large a swarm can grow before contacts appear. The task decides how many contacts it has at any given size.

\subsection{Computational Performance and Scalability}
\label{sec:performance}
The experiments run one task per core on shared compute nodes, so other jobs on the same
node could slow down the solve time. We repeated the measurement on an idle node reserved for
this run alone. Averaged over the density sweep, each formulation's mean agrees with its
shared-node value to within $0.07\,\mathrm{ms}$, so sharing a node does not distort the
per-robot times reported in Section~\ref{sec:comparison}, and the idle run also gives the
full distribution. Each solve covers the full per-robot control step, including both
consensus iterations.

The mean and the $99$th percentile stay well inside both the $100\,\mathrm{ms}$
($10\,\mathrm{Hz}$) and $33.3\,\mathrm{ms}$ ($30\,\mathrm{Hz}$) budgets at every density.
Across the six formulations $p_{99}$ peaks at $13.1\,\mathrm{ms}$, and at the operating
densities $N_b\ge100$ it stays at or below $3.9\,\mathrm{ms}$. Worst-case single solves are
larger and matter for a hard real-time guarantee. The maximum observed on the idle node is
$98.8\,\mathrm{ms}$, so the $10\,\mathrm{Hz}$ budget is met at every density while the
$30\,\mathrm{Hz}$ budget is not.
These maxima are lone draws from the tail. They grow with $N_b$ only because a larger swarm
performs more solves per run, from $3000$ at $N_b=10$ to $41100$ at $N_b=137$, while the
mean and $p_{99}$ do not grow. A $30\,\mathrm{Hz}$ deployment would therefore cap the
solver iterations and fall back to the reactive controller on the rare overrun.

The per-\emph{step} wall time necessarily grows with $N_b$, since a single process advances
all robots in sequence. It is at least $N_b$ times the per-robot cost, so
$\gtrsim 37$, $93$, $184$ and $374\,\mathrm{ms}$ at $N_b=10,25,50,100$. This is a simulation
artifact. Each robot runs its own optimization onboard or in parallel when deployed, so the
governing cost is the per-robot time. That cost is flat in $N_b$
(Section~\ref{sec:comparison}), so computation does not limit the swarm size. The current single-process simulation bottleneck can be removed through parallel execution and is left as future implementation work.

\section{Conclusion}
\label{sec:conclusion}

We presented a distributed predictive flocking controller for dense miniature ellipsoid robot swarms in a confined arena. Each robot solves a local NMPC combining task tracking, Reynolds-style flocking, and inter-robot and obstacle safety constraints, using only exchanged predicted trajectories of active neighbors, without a leader, central planner, or global map. The results demonstrate that the same controller architecture generalizes across different modal objectives such as gradient-ascending phototaxis, stable phototactic orbital formation, and even flocking in the absence of any stimulus. The emergent collective behavior arises purely from local optimization of cohesion and alignment costs, coupled with safety constraints that maintain separation. This separability between the task and the coordination points towards a modeling paradigm where new tasks can plausibly be added without necessarily re-deriving the safety features or coordination structure. Each robot optimizes against a fixed neighbor set, so solve time is $\mathcal{O}(1)$ in swarm size, measured at $3.7\,\mathrm{ms}$ on average and flat across the density sweep, making the scaling limit architectural (a serial solve with a single workstation caps closed-loop control at roughly $27$ robots) rather than algorithmic.

Comparing six safety formulations under identical conditions showed that performance is determined by structure rather than barrier order; pairing a slacked barrier anchor with a rate condition reduces contact counts by a factor of $2.4$ relative to formulations without an anchor, with no overlap across the density sweep. Considering the phototaxis mode as the deployment target, the LA Slack Dt-CBF is found to be most effective in preventing collisions, while also providing the highest inter-robot margins for most swarm sizes tested. The look-ahead handle drives this improvement, giving the turn rate direct authority over the constraint so a robot can steer around a neighbor rather than only brake. Removing this handle alone raises contacts by up to a factor of $2.5$ at low density, and by $1.2$ at the geometric packing limit. Moreover, a higher-order rate chain improves performance further at moderate density but loses its advantage at the packing limit, showing that a higher barrier order does not automatically guarantee better safety.

The collective motion is analyzed through local polar order and NND analysis, which, apart from quantifying controller performance, also point toward a biomimetic reading of these results. The muted order decay and more efficient packing found for the orbital mode, versus the sharper degradation under phototaxis, mirror a structural distinction well documented in natural swarms: neighbors on a shared circulating path align through a shear-like velocity field, while neighbors converging on a common point align through a convergent one, with the latter being inherently more collision-prone and alignment-averse as group size grows. That our controller reproduces these qualitative results from local optimization alone, without it being an explicit design target, suggests the polar order and NND metrics already computed here could serve as a bridge to the natural flocking literature. Furthermore, the wall-following behavior observed in the absence of external stimuli concurs well with studies of natural fish schools and insect swarms in confined spaces, a systematic evaluation and comparison of which can open a path to tuning the flocking and safety weights toward biologically observed regimes rather than controller-internal ones.

The current work has two limitations. The vibration and friction induced locomotion of the real bristlebot is modeled only kinematically rather than physically, so variations involving surface, payload, and actuation asymmetry enter the controller only as unmodeled disturbance. The safety comparison is also internal by design: six formulations within one stack, which isolates the safety constraint but does not benchmark against external distributed-flocking baselines, most of which target fully actuated 3D quadrotors or underactuated 2D robots rather than vibration-driven platforms. Further extensions can be carried out in terms of optimization and learning; the slack weight, rate-condition horizon, and look-ahead distance form a compact design space suited to automated tuning against the contact-rate and feasibility metrics already collected here, while a learned residual dynamics model could absorb the friction disturbance currently left unmodeled, narrowing the gap between the kinematic model and the true vibration-driven actuation. Immediate next steps include physical validation on the GRASPion hardware, extension to dynamic obstacle avoidance, and robust or stochastic MPC formulations for stronger safety guarantees. Overall, these results establish a single, kinematic model-agnostic NMPC architecture that unifies task-driven behavior, emergent flocking, and safety against collisions, while the polar-order and NND signatures it reproduces open a concrete path toward grounding swarm-robotics control in the structure of natural collective motion.

  \printcredits

  \section*{Declaration of competing interest}
  The authors declare that they have no known competing financial interests or personal
  relationships that could have appeared to influence the work reported in this paper.
  
  \section*{Data availability}
  No external data was used. The code and simulation environment that reproduce all results are available at \url{https://github.com/prajolshrestha/bristlebot-swarm-mpc}.
  The full set of simulation videos, including the full-length renders from which the supplementary videos are derived, is archived at
  \url{https://doi.org/10.5281/zenodo.22642863}.


  \bibliographystyle{unsrtnat}

  \bibliography{cas-refs}
  
\newpage
\appendix
\section{Collective behaviors with static obstacles}
\label{app:behaviors_obstacles}

\begin{figure}[pos=htbp]
  \centering
  \includegraphics[width=\linewidth]{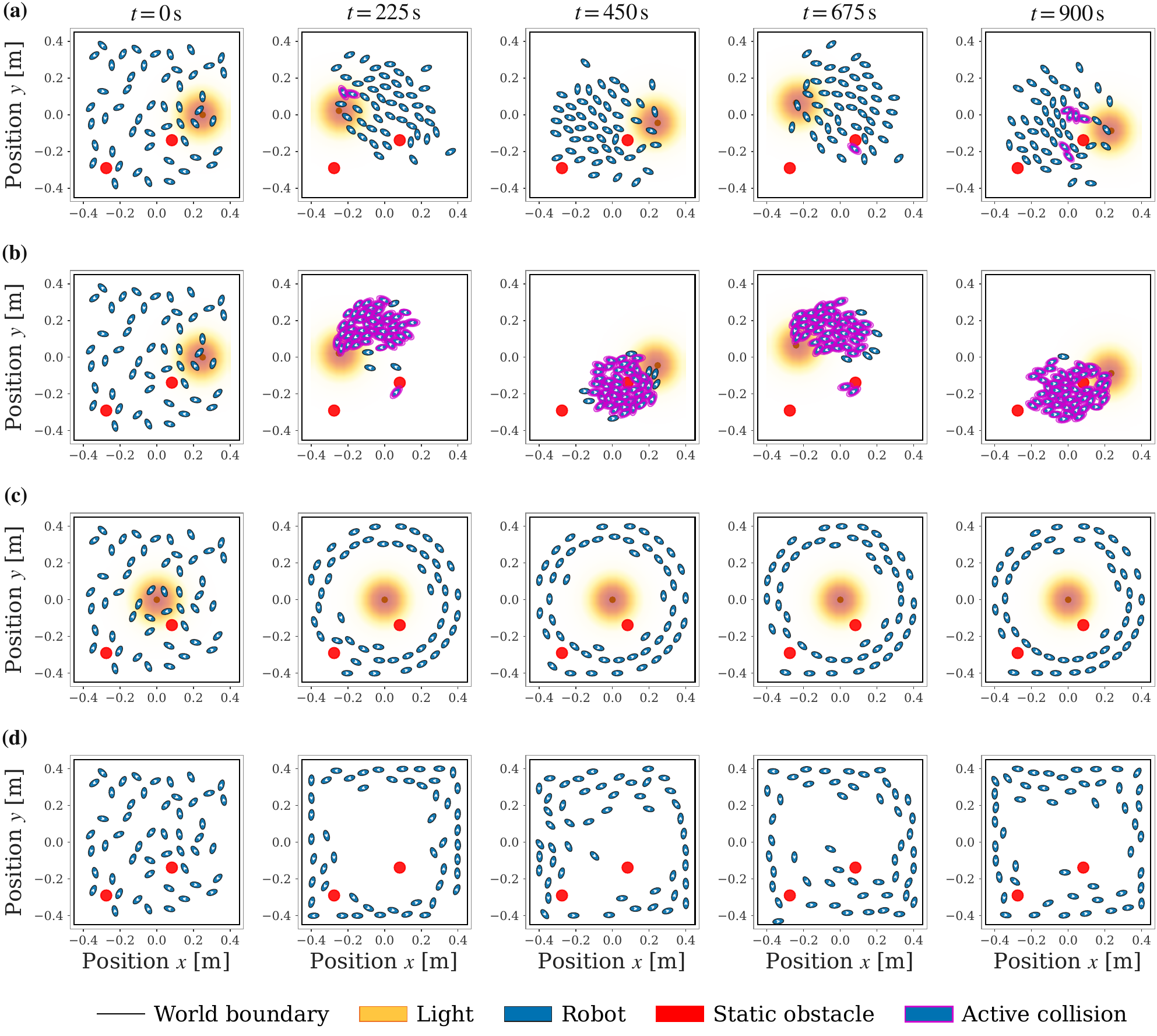}
  \caption{Collective behaviors with two static obstacles (LA Slack Dt-CBF formulation,
  $N_b=50$), one behavior per row. (a) Phototaxis, (b) phototaxis with the Reynolds coupling
  removed, (c) phototactic orbiting, and (d) cohesion without a stimulus. Columns are snapshots at $t=0$, $225$, $450$, $675$ and $900\,\mathrm{s}$. The red disks are the static obstacles and the orange
  field is the light source. Supplementary video~2 shows these four behaviors in the two-obstacle arena.}
  \label{fig:behaviors_2obs}
\end{figure}

\begin{figure}[pos=htbp]
  \centering
  \includegraphics[width=\linewidth]{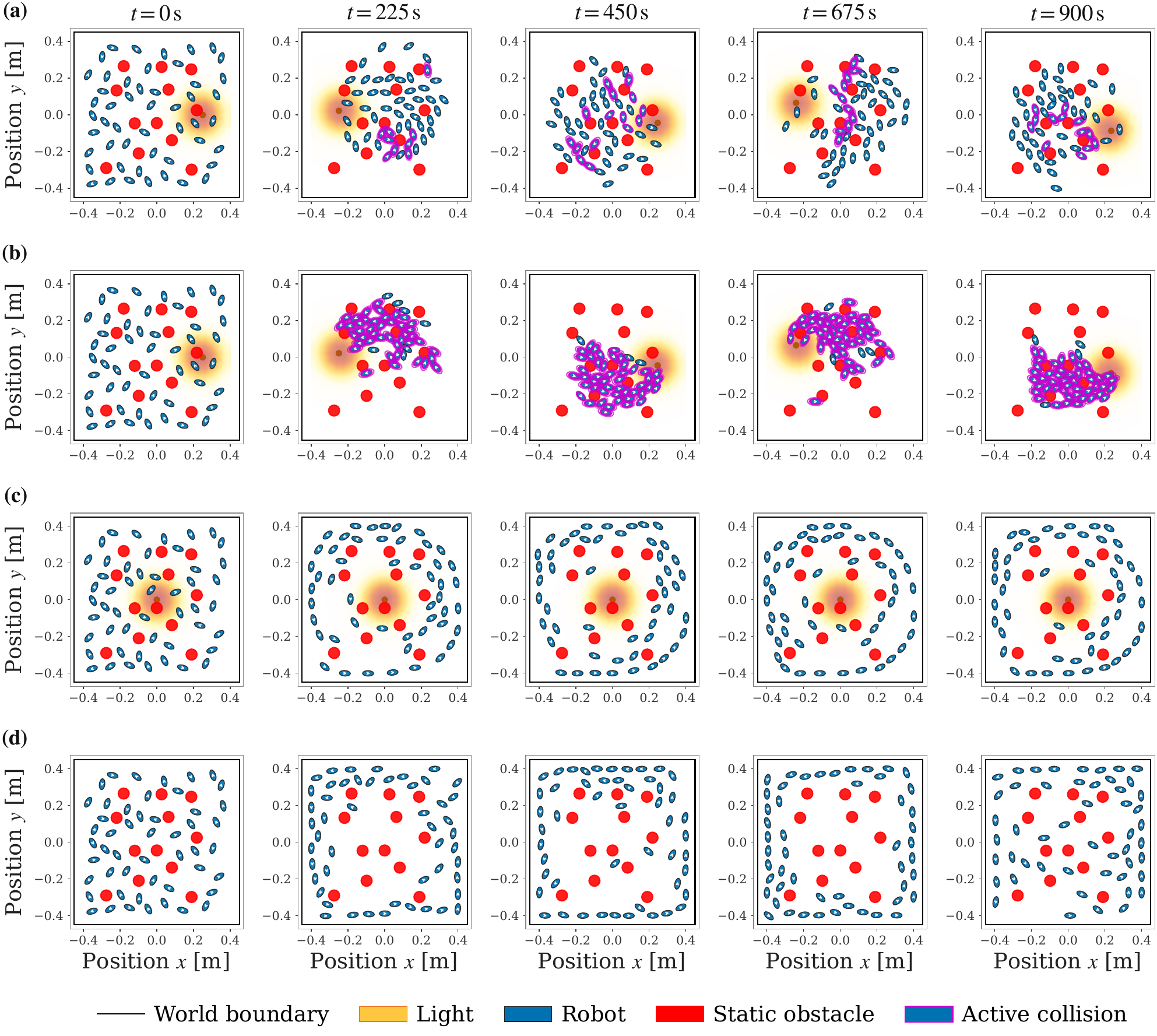}
  \caption{Collective behaviors with twelve static obstacles (LA Slack Dt-CBF formulation,
  $N_b=50$), one behavior per row. (a) Phototaxis, (b) phototaxis with the Reynolds coupling
  removed, (c) phototactic orbiting, and (d) cohesion without a stimulus. Columns are snapshots at $t=0$, $225$, $450$, $675$ and $900\,\mathrm{s}$. The red disks are the static obstacles and the orange
  field is the light source. Supplementary video~3 shows these four behaviors in the cluttered arena.}
  \label{fig:behaviors_12obs}
\end{figure}

Figs.~\ref{fig:behaviors_2obs} and~\ref{fig:behaviors_12obs} repeat the four panels
of Fig.~\ref{fig:behaviors} in cluttered arenas, with two and with twelve static
obstacles. Each mode keeps the same qualitative behavior while the controller routes
the swarm around the obstacles.

\clearpage
\section{Supplementary data}
\label{app:supplementary_data}

Table~\ref{tab:params} lists the complete set of controller, safety, environment, and
simulation parameters used to produce every result, so the setup of
Section~\ref{sec:simulation} is reproducible from the paper alone.

\begin{table}[pos=htbp]
  \centering
  \small
  \caption{Complete controller, safety, environment, and simulation parameters used for
  all experiments. Robot-scale lengths are in cm, the arena and sensing ranges in m, and
  speeds in m/s; cost and slack weights are dimensionless. Values are those of the deployed
  LA Slack Dt-CBF; the cost weights, horizon, arena and platform limits are shared by
  every formulation, while $\alpha$, $L$, $\gamma_1,\gamma_2$ and the slack weights are
  re-tuned per formulation (Section~\ref{sec:sim_setup}). The
  angular-rate bounds are asymmetric, reflecting the vibration-driven platform's differing
  left- and right-turn authority.}
  \label{tab:params}
  \begin{tabular}{ll}
    \toprule
    \textbf{Parameter} & \textbf{Value} \\
    \midrule
    \multicolumn{2}{l}{\emph{Model and actuation bounds}}\\
    Arena $\mathcal{A}$                              & $0.8\,\mathrm{m}\times0.8\,\mathrm{m}$ ($0.64\,\mathrm{m^2}$) \\
    Position bounds $p_x,p_y$                         & $[-0.4,\,0.4]\,\mathrm{m}$ \\
    Linear speed $v$                                  & $[0.01,\,0.15]\,\mathrm{m/s}$ \\
    Angular rate $\omega$                             & $[-3.463,\,1.432]\,\mathrm{rad/s}$ \\
    Robot semi-axes $a,\,b$                           & $3\,\mathrm{cm},\ 1.5\,\mathrm{cm}$ \\
    \addlinespace
    \multicolumn{2}{l}{\emph{Cost weights (dimensionless)}}\\
    Position, heading $q_{\rm pos},q_{\theta}$        & $60,\ 80$ \\
    Speed, turn rate $q_v,q_{\omega}$                 & $1.5,\ 1.5$ \\
    Control effort $r_u$                              & $0.01$ \\
    Cohesion, alignment $q_{\rm coh},q_{\rm ali}$     & $2.0,\ 5.0$ \\
    Slack quadratic $Z^{\rm rob}\!=\!Z^{\rm obs}$     & $10^{8}$ \\
    Slack linear $z^{\rm rob}\!=\!z^{\rm obs}$        & $10^{5}$ \\
    \addlinespace
    \multicolumn{2}{l}{\emph{Safety and barrier-CBF}}\\
    Safe distance $D_{\rm safe}\!=\!D_{\rm safe,obs}$ & $7\,\mathrm{cm}$ \\
    Decay $\alpha$ (inter-robot)                      & $0.9$ \\
    Decay $\alpha_{\rm obs}$ (obstacle)               & $0.9$ \\
    Look-ahead handle $L$                             & $2.1\,\mathrm{cm}$ \\
    Dt-HOCBF gains $\gamma_1,\gamma_2$                & $0.5,\ 0.5$ \\
    Dt-HOCBF obstacle gains $\gamma_{1,\rm obs},\gamma_{2,\rm obs}$ & $0.5,\ 0.5$ \\
    \addlinespace
    \multicolumn{2}{l}{\emph{Communication, sensing, environment}}\\
    Communication radius $R_{\rm comm}$               & $0.15\,\mathrm{m}$ \\
    Max neighbors, obstacle slots $K,M$              & $5,\ 5$ \\
    Consensus iterations $n_{\rm cons}$               & $2$ \\
    Obstacle sensing range $d_{\rm obs}$              & $0.30\,\mathrm{m}$ \\
    Obstacle radius $r_m$                             & $3\,\mathrm{cm}$ \\
    Obstacle counts (free/few/cluttered)             & $0\,/\,2\,/\,12$ \\
    Light spread $\sigma_{\rm L}$, orbit radius $r$           & $0.10\,\mathrm{m}$, $0.25\,\mathrm{m}$ \\
    Source speed $v_s$                           & $0.15\,\mathrm{m/s}$ \\
    Behavior look-ahead $d_{\rm la}$, ref.\ speed $v^{\rm ref}$ & $7\,\mathrm{cm}$, $0.10\,\mathrm{m/s}$ \\
    \addlinespace
    \multicolumn{2}{l}{\emph{Simulation and evaluation}}\\
    Horizon $N$, control step $\Delta t$             & $20$, $0.1\,\mathrm{s}$ \\
    Episode length                                    & $9000$ steps ($900\,\mathrm{s}$) \\
    Random seeds                                       & $10$ \\
    Integrator                                         & ERK-4, 3 steps/interval \\
    QP solver                                          & acados SQP-RTI, HPIPM, Gauss-Newton \\
    \bottomrule
  \end{tabular}
\end{table}


\end{document}